# CKMorph: A Comprehensive Morphological Analyzer for Central Kurdish

Morteza Naserzade
*Sharif University of Technology*
morteza.naserzade1983@gmail.com

Aso Mahmudi
*University of Tehran*
aso.mahmudi@ut.ac.ir
orcid: 0000-0003-2155-4101

Hadi Veisi*
*University of Tehran*
h.veisi@ut.ac.ir
orcid: 0000-0003-2372-7969

Hawre Hosseini
*Ryerson University*
hawre.hosseini@ryerson.ca

Mohammad MohammadAmini
*Avignon University*
mohammad.mohammadamini@univ-avignon.fr

2021

**Abstract**

A morphological analyzer, which is a significant component of many natural language processing applications especially for morphologically rich languages, divides an input word into all its composing morphemes and identifies their morphological roles. In this paper, we introduce a comprehensive morphological analyzer for Central Kurdish (CK), a low-resourced language with a rich morphology. Building upon the limited existing literature, we first assembled and systematically categorized a comprehensive collection of the morphological and morphophonological rules of the language. Additionally, we collected and manually labeled a generative lexicon containing nearly 10,000 verb, noun and adjective stems, named entities, and other types of word stems. We used these rule sets and resources to implement *CKMorph Analyzer* based on finite-state transducers. In order to provide a benchmark for future research, we collected, manually labeled, and publicly shared test sets for evaluating accuracy and coverage of the analyzer. CKMorph was able to correctly analyze 95.9% of the first test set, containing 1,000 CK words morphologically analyzed according to the context. Moreover, CKMorph gave at least one analysis for 95.5% of 4.22M CK tokens of the second test set. The demonstration of the application and resources including CK verb database and test sets are openly accessible at https://github.com/CKMorph.

## 1 Introduction

Morphology is a branch of linguistics that deals with the study of rule-based and regular changes in the form and meaning of words and determines how morphemes are combined in compound words (Haspelmath & Sims, 2013). A typical morphological analyzer uses morphological rules and language regularities to analyze words with complicated structure and break them down into their comprising morphemes while determining each morpheme's role in the word structure. Morphological analyzers, especially in morphologically complex languages, are a critical component in many other tools and downstream applications, such as stemmers and lemmatizers, spell checkers , machine translation, part of speech taggers , and so forth (Lindén et al., 2011). For many languages, morphological analyzers have been developed (Çöltekin, 2010; Megerdoomian, 2004; Schmid et al., 2004) including for under-resourced languages such as Iñupiaq (Bills et al., 2010) and Evenki (Zueva et al., 2020). However, no robust and comprehensive morphological analyzer has yet been developed for the Kurdish language. The reasons for this, as elaborated on in the paper, are rich morphology, lack of generative lexicons, and other data such as verbs database, among others. In this study, while manually and automatically labeling necessary datasets and lexicons (publicly shared), we collect a comprehensive set of morphological rules of Central Kurdish (CK) group (ISO 639-3: ckb) and implement a two-level morphological analyzer on that basis.

Three requirements of a morphological analyzer are: 1) *lexicon*, containing a list of the word stems of the language tagged by the their morphological class; 2) *morphotactics*, a model for ordering the morphemes inside the words of the language; and, 3) *orthographic rules*, a model for applying the changes that occur in orthography of the word when the morphemes combine (Jurafsky & Martin, 2014). A widely used technique in computational morphology is a two-level approach to morphology analysis where there are two levels of surface and lexical. At the surface level, a word is represented in its original orthographic form. At the lexical level, a word is represented by denoting all of its functional components. A two-level morphology analyzer is able to perform two tasks: analysis and generation. As to "analysis", in the field of computational morphology, it refers to the process of getting an underlying morphological description from the surface form, e.g., analysis of the Kurdish word */dengim/* 'my voice' gives "*/deng/*[Noun]<+1st><+Singular>". The "generation", on the other hand, is the inverse process. Two-level morphology descriptions are usually implemented using Finite-State Transducers (FSTs) (Lindén et al., 2011).

Our study has two main parts. First, we present the outline of the lexicon collection and a comprehensive report on the morphotactics and orthographic rules of CK corresponding to the requirements of the morphological analyzers as described earlier. In the second part, we describe the steps of the process of implementing our proposed two-level morphological analyzer for CK with the help of the various data sources we have collected and labeled. The implementation is based on Helsinki Finite-State Technology (HFST) (Lindén et al., 2011).

It is worthy of mentioning that Kurdish is an umbrella term for several dialect groups including Northern, Central, Southern, and Zaza/Gorani among others, with millions of speakers. Implementing a single morphological analyzer for all Kurdish dialects spoken in several countries is not a practical task. The reason is that each dialect has its own set of phonological, lexical, morphological, and syntactic variations, gathering all of which requires considerably large amount of time and resources. Moreover, the Northern dialects usually use Latin script, and the others use Arabic script with different standards for each dialect group. Therefore, in this paper, we introduce a two-level morphological analyzer for CK group, and we focus mainly on the forms and structures used in the official texts known as Standard Central Kurdish (SCK). This study can serve as a stepping stone to implement morphological analyzers for other Kurdish dialects.

Throughout this paper, we discuss the main challenges we faced for implementation of CKMorph, which are mentioned in short here:

1. *Rich morphology*: Implementing the complicated morphotactic rules of a synthetic language as CK needs hard working as the language has features like variety of verb types, ergativity, semantic restrictions, far dependency of bound morphemes, displacement of clitics, and generative compound structures. To face this challenge, we not only implemented an extensive set of inflectional rules, but we also complemented them with some derivational rules; in this way, we ensured that we achieve considerable coverage using a relatively limited lexicon.

2. *Lack of a generative lexicon*: As CK is a low-resource language, we spent considerable time collecting and manually labeling the lexicon. We had to manually extract and categorize the word stems from the few available old-fashioned dictionaries and the newly coined words and named entities from raw news corpora. The reslting lexicon contains over 10,000 nouns, adjectives, named entities and other types of word stems.

3. *Lack of a verb database*: Analysis of verbs is specifically challenging due to various aspects with respect to which they should be analyzed. Especially in CK morphology, verbs can form richer morphological structures. For instance, due to ergativity feature, the transitive past verbs accept different pronoun sets. We, for the first time, prepared a comprehensive database of CK verbs. We gathered the past, present, and passive/causative stems, transitivity, valence, valid adverbial affixes, and semantic restriction for more than 400 verb lemmas. These features are necessary for correct analysis of the verbs.

We believe that this study can serve two main purposes. Firstly, it can be used as a manual for CK computational morphology. Secondly, our study can act as a benchmark for future studies, and the resources and the test set provided can be leveraged to foster research in this area. The main contributions of this research can be enumerated as follows:

- Providing a comprehensive morphotactic (morphological) rule set for all types of words in CK, complemented by some derivational rules for the sake of a more generative lexicon and more coverage;
- Providing a complete orthographic (morphophonemic) rule set for all types of words in CK;
- Providing a comprehensive database of CK verbs including their inflectional properties and valences;
- Providing a test set for benchmarking purposes containing 1000 words in context;
- Implementing a CK morphological analyzer with state of the art performance.

The rest of this document is organized as follows: In section 2, we present a background about Central Kurdish and its orthographic system and the challenges for its morphological analysis, along with a review of previous work in this area. In section 3, we give details about the collecting our lexicon. In section 4, CK morphotactics are discussed thoroughly. Orthographic rules of CK are described in section 5. In section 6, the implementation steps are presented. In section 7, the proposed analyzer is evaluated and its shortcomings are discussed. Section 8 is devoted to a summary of our research and suggestions for directions of future work in this area.

## 2 Background

### 2.1 Kurdish and its Morphological Analysis Challenges

Kurdish is an umbrella term for several dialects spoken by millions of Kurds in the middle east. Speakers of Central Kurdish (CK), a major dialect group of Kurdish, mostly reside in western Iran and northeastern Iraq. CK has several sub-dialects such as Sulaimani, Mukri, Erbili, Sineyi, Jaffi/Garmyani which are slightly different in phonology, morphology and syntax.

Although there is not officially determined which dialect of Kurdish is the official language in Kurdistan Regional Government and in Iraq, CK is used in majority of state documents. The speakers of CK in Iraq and Iran use a special alphabet based on Arabic script. However, in this study, for greater readability in describing the morphotactics, we use a slightly modified version of Standard Latin Alphabet of Northern Kurdish, which is in use mainly in Turkey and Syria. The conversion from our phonemic system to the CK alphabet is straightforward and can be done using Table 1. Note that in three cases, there is no one-to-one mapping between the orthographical letters and the phonemes of the language (Mahmudi & Veisi, 2021). The phonemes /u/, /w/, and /û/ are represented by letter و and phonemes /y/ and /î/ by letter ى. The short vowel /i/ is not represented in the CK alphabet, e.g., the word /genim/ 'wheat' is written as "گەنم".

| This Study: | /ʔ/ | /a/ | /b/ | /p/ | /t/ | /c/ | /ç/ | /ḧ/ | /x/ | /d/ | /r/ | /ř/ | /z/ | /j/ | /s/ | /ş/ | /ɛ/ | /ẍ/ | /f/ |
|---|---|---|---|---|---|---|---|---|---|---|---|---|---|---|---|---|---|---|---|
| IPA: | ʔ | äː | b | p | t | dʒ | tʃ | ħ | x | d | ɾ | r | z | ʒ | s | ʃ | ʕ | ɣ | f |
| CK alphabet: | ئ | ا | ب | پ | ت | ج | چ | ح | خ | د | ر | ڕ | ز | ژ | س | ش | ع | غ | ف |
| This Study: | /v/ | /q/ | /k/ | /g/ | /l/ | /ł/ | /m/ | /n/ | /w/ | /u/ | /û/ | /o/ | /h/ | /e/ | /y/ | /î/ | /ê/ | /i/ | |
| IPA: | v | q | k | g | l | ɫ | m | n | w | ʊ | uː | oː | h | a | j | iː | ɛː | ɪ | |
| CK alphabet: | ڤ | ق | ک | گ | ل | ڵ | م | ن | و | و | وو | ۆ | ھ | ە | ى | ى | ێ | | |

Table 1: Phonemic writing system of this study compared to Central Kurdish (CK) alphabet and IPA.

The diversity of the dialects is a challenge especially in the computational morphology field. As the bound morphemes and their order are different in dialects of CK, implementing a single tool capable of morphologically analyzing inputs from all dialects will be a laborious task. For example, the present perfect verb /hatuwete/ 'he has came to' is conjugated differently in various dialects such as /hatote/, /hatuwese/, /hatîyete/, /hatîte/, /hatîyese/, /hatigese/, /hatete/ and /hatese/. In this study, we describe mainly the Standard Central Kurdish (SCK), a form of the language that is used in the official and educational textual documents by Kurdish speakers in Iraq and Iran. SCK has developed on the basis of the dialect of Sulaimani in Iraq, where since 1918, has been the official language of the local administration and of primary education is Kurdish (Wahby & Edmonds, 1966, Introduction). However, we have endeavored to include some of the morphological structures from other CK dialects that are commonly seen in the written texts.

Another major challenge is with respect to the synthetic nature of CK, i.e., its words are usually made up from more than one morpheme. For example, the word /kuřekanîşmanin/ is made up from noun stem /kuř/ and five bound morphemes: /kuř-ek[e]-an-îş-man-in/

(1) kuř-ek[e]-an=îş=man=in
    son-DEF-PL=ALSO=1PL.POSS=3PL.COP

    'they are also our sons'

or the verb /netandegirtînewe/ is made up from verb stem /girt/ and also five bound morphemes:

(2) ne-tan-de-girt-în=ewe
    NEG-3PL-IMPERF-catch-1PL=AGAIN

    'you were not catching us'

Considering the possible combination of suffixes and enclitics that a noun can take in CK (see Figure 1), we can generate 2+8×(2+2×7×(7+3×6)) = 2818 well-formed words from a single noun stem. This number is higher for the verbs, especially transitive ones. Further, some special morphological features of CK such as "ergativity" and "displacement of clitics" have led to more complexity and challenges in language processing and especially morphological analysis. Additionally, in some cases, there are more than one output for a specific word. For example, suffix /-im/ in the word /kuř-im/ has four analyses as shown in the following sentences:

- Copula: /min kuř-im/ 'I am a boy'
- Possessive pronoun: /ʔew kese kuř-im nîye/ 'that person is not my son'.
- Subject pronoun: /kuř-im nebînî/ 'I did not see (a) boy'
- Object/Complement pronoun: /kuř-im pê nîşan bide/ 'show me (a) boy'

## 2.2 Related Works

There are numerous descriptive studies about the grammar of the Kurdish dialects. Among such works, we have extensively used those that have provided relatively more practical linguistic resources that can be used for computational purposes, for instance, in order to build rule sets. Some of the most prominent works of the aforementioned type are Amin (2016), MacKenzie (1961), McCarus (1958), and Walther (2012). These works have endeavored to formulate the morphological rules of CK. Building on such works, we have collected and uniformly translated such rules for use in computational settings.

A major challenge of the Kurdish language processing is the lack of lexical resources and corpora. Additionally, majority of the existing Kurdish dictionaries are not compiled using lexicographical standards. They are not tagged with part-of-speech and they contain entries from various dialects. More importantly for the natural language processing (NLP) tasks, they are usually published in printed form and only a few are digitally available. Therefore, extracting a trustworthy categorized list of word-stems is a time-consuming task. Walther and Sagot (2010) suggest a three-step method to develop a morphological lexicon for CK within a framework called Alexina. They collect a small raw corpus with 591K tokens and 63K unique tokens from which they extract a lexicon named SoraLex containing 17.6K words. Further, they collect 68 verb roots of the Kurdish language and assert that this collection includes 25% of Kurdish verbs. Veisi et al. (2019) introduce the AsoSoft text corpus as the first large-scale text corpus for the CK containing 188M tokens, mainly collected from online news agencies, books and magazines. We use this text corpus for various tasks including evaluation purposes and selection of standard variations, as described later in the paper. Ahmadi et al. (2019) mention the lack of machine-readable lexical resources and present lexicons in the standard ontology for Central, Northern and Hawrami Kurdish. Their CK lexicon contains 5.4K lexemes extracted from "Sorani Kurdish: A Reference Grammar with Selected Readings" by W. M. Thackston. Due to lack of part-of-speech tags, we were not able to leverage this resource.

Here we review the few studies in the field of computational morphology of the Kurdish language. It is to be noted that the rule-based morphological analyzers are generally divided into "affix strippers" and "two-level morphology based analyzers" (Jurafsky & Martin, 2014). The affix strippers decompose the word into its smallest possible morphemes; i.e., considering the spelling and morphotactics rules, they remove the affixes to find a valid stem; if the stem belongs to the lexicon, the word will be identified as a well-formed word and analyzed, otherwise the word is identified as ill-formed and will be discarded. The affix strippers are usually used for languages with less complex morphology in fast methods of stemming, lemmatizing, and information extraction (Jurafsky & Martin, 2014). The computational morphotactics of CK is mentioned first in Baban and Husein (1995) where the general structure of Central Kurdish verbs is formalized. Furthermore, the CK morphotactics is discussed in the contexts of the NLP task such as stemming and information retrieval. Saeed, Rashid, Mustafa, Agha, et al. (2018), Saeed, Rashid, Mustafa, Fattah, et al. (2018), Salavati (2013), and Salavati et al. (2013) list common inflectional and derivational affixes and verb structures of CK used in rule-based affix-stripping stemmers for improving the Kurdish text classification and information retrieval tasks. Hosseini et al. (2015) use a semi-supervised method to produce a generative CK lexicon (KSLexicon) containing 35K words. Also, they try to extract and formulate morphological rules of the verbs, nouns, adjectives and adverbs of CK.

The two-level morphology was introduced by Koskenniemi (1983). This method maps the lexical level (functional components of the word) to the surface level (a well-formed word) considering the phonetic and orthographic rules of the language. The main improvement of the two-level morphology is applying the rules in parallel, rather than in a cascading manner. Therefore, it is efficient for analysis and generation of synthetic and morphologically complex languages. In the two-level morphological analysis, words are regenerated, morpheme by morpheme, based on the valid morphological rules of the language. With addition of each morpheme, some information about the structure of the word will be added into the lexical level. The compilers and implementation methods of the two-level use "finite-state transducers (FSTs)"; i.e., the morphotactics which determine the proper sequencing of the morphemes are encoded as FSTs. For more details about the two-level morphology and its implementation using FSTs one can refer to (Beesley & Karttunen, 2003).

Two-level morphology has been used in morphological analyzers of numerous syntactic languages such as Finnish (Koskenniemi, 1983, Chapter 3), Turkish (Oflazer, 1995), and Plains Cree (Harrigan et al., 2017). Computational morphology of the Persian language which is an Iranian language closely related to CK is studied in detail (Megerdoomian, 2000; Riazati, 1997). Megerdoomian (2004) introduces a finite-state morphological analyzer of Persian dealing with similar challenges like long-distance dependencies. Heidarpour et al. (2021) introduces a more complete Persian morphological analyzer including unofficial and conversational texts. For CK, Ahmadi and Hassani (2020) present a brief sketch of CK morphological rules and basic instructions for implementing a finite-State transducer based on tools developed by Schmid et al. (2004) as the experimental environment. Compared to these previous works on CK morphotactics, we have extended the elaborated rules for all word classes. We have also prepared a more generative lexicon that covers more textual content of CK.

# 3 Lexicon

A primary requirement of a morphological analyzer is a lexicon, containing a list of word stems tagged by their morphotactic category. For a practical application, the lexicon must cover a wide range including basic and frequent, classical and modern lexemes, as well as named entities. The process of preparing a generative lexicon is time-consuming, especially for under-resourced languages.

Dictionaries are the fundamental resources for compiling a lexicon. However, most Kurdish dictionaries, such as "Henbane Borîne" (Kurdish-Persian, 57.8K entries) (Sharafkandi, 1990) and "Ferhengî Xał" (Kurdish, 31.5K entries) (Khal, 2000), are not efficient and trustworthy for computational tasks since they do not have part-of-speech tags and their entries are a mixture of different dialects without accurate dialect tags. The pioneer standard "A Kurdish-English Dictionary" (Wahby & Edmonds, 1966) with about 7K main entries and 18.9K sub-entries is a concise and accurate resource for collecting an essential CK lexicon. The entries of this dictionary have tags for part-of-speech (including transitivity of the verbs). The main advantage of this dictionary is that it mainly includes entries from the dialect of Sulaimani, which is very close to the current SCK.

The most significant part of our lexicon is a comprehensive database of CK verbs. The inflection of the CK verbs, as described in Section 4, is more complex than other CK's word classes. For faultless implementation of morphotactic rules, we should know the properties of every single verb lemma. The following examples show some of the features that require special attention to the CK verbs:

- The onomatopoeic verbs (such as */qiřandin/* 'to scream') are intransitive semantically; however, they are inclined as transitive verb with no objects.

- The occurrence of all adverbial prefixes (such as /heł, da, wer, .../) which can change the meaning of the verbal structures, are not allowed with all verbs. For instance, the prefix /wer-/ is valid with the verb /girtin/ 'to take' and yields /wergirtin/ 'receive', but /wer-/ is not allowed before /zanîn/ 'to know' (/*werzanîn/).
- The passive stem for some transitive verbs such as /wîstin/ 'to want' is irregular (/wîstr-/).
- Some transitive verbs like /gutin/ 'to say' only accept third person objects.
- Some intransitive verbs such as /kewtin/ 'to fall' do not make causative structures.

For more than 400 verb lemmas, we have collected the stems (past, present, and passive/causative), transitivity, valence, valid adverbial affixes, and semantic restriction. Furthermore, we have collected more than 1300 verbal compounds which in turn, can generate many more compound nouns and adjectives. Our main resources for collecting the CK verb database were "A Kurdish-English Dictionary" by Wahby and Edmonds (1966) and the appendix of the book "Kurdish Dialect Study" by MacKenzie (1961), where details about all CK verbs are given. Some errors and inconsistencies were seen in some of the verbs, which were corrected by checking the other dictionaries and corpora.

The general Kurdish dictionaries usually lack the newly coined words of the language (including technology and sport-related terms) and proper names and other named entities. Therefore, we also need text corpora, especially recently created texts like news websites. Over recent years, there have been attempts in order to develop text corpora for the Kurdish language (Hosseini et al., 2015; Sheykh Esmaili et al., 2013; Veisi et al., 2019; Walther & Sagot, 2010). We used the AsoSoft Central Kurdish text corpus (Veisi et al., 2019) which was the largest corpus at the time and contains almost 188M words. First, we extracted the unique words of the corpus sorted by frequency and examined them by our basic analyzer with lexicon made from the dictionary entries. Then words that were not detected were checked manually for the new stems. Each word stem was proofread before being added to the lexicon. If a stem had more than one variation, especially in the case of foreign loanwords or named entities, we chose one variation as the main entry and the other tagged as non-standard variations, based on the standardization rules and the corpus frequency (Mohammadamini et al., 2019).

Table 2 shows the statistics of lexems added to the lexicon.

| Category | Example | Count |
| --- | --- | --- |
| intransitive lemmas | کەوتن | 308 |
| transitive lemmas | خستن | 125 |
| onomatopoeic lemmas | قیژ | 66 |
| compound lemmas | ئابڕوو+بردن | 1220 |
| general nouns | ئاو، زەوی | 4000 |
| units | کیلۆ | 44 |
| titles | دادە، مەلا | 36 |
| definite nouns | جەناب، پەنجاکان | 15 |
| person's proper names | نالی | 955 |
| location's proper names | سلێمانی | 530 |
| organization's proper names | پێشمەرگە | 198 |
| date's proper names | شەممە، ئەیلوول | 51 |
| ethnic group's proper names | کورد | 48 |
| demonym's proper names | کوردستانی | 47 |
| language's proper names | کوردی | 42 |
| miscellaneous proper names | بیتکۆین، کۆرۆنا | 67 |
| general adjectives | ڕوون | 1200 |
| non-gradable adjectives | ون، هەتاوی | 93 |
| indeclinable adjectives | ئەو، چەند | 13 |
| havable adjectives | برسی | 6 |
| adverbs | بڕایانە | 364 |
| numerals | پەنجا | 42 |
| personal pronouns | من | 6 |
| special pronouns | خۆ، یەکتر | 28 |
| indeclinable prepositions | بەرەو | 45 |
| prepositions getting /-da/ | بەگژ | 28 |
| prepositions getting /-da/ and /-ewe/ | لەنزیک | 18 |
| prepositions getting /-ewe/ | بەپیر | 12 |
| interjection | ئافەرین | 33 |
| subordinating conjunctions | ئەگەر | 35 |
| coordinating conjunctions | یان | 8 |
| contraction | لێرە | 25 |
| particle | نا، بەڵێ | 10 |
| letters | جیم، ئێف | 59 |
| abbreviation | د.، هتد | 3 |
| **Total** | | **9,780** |

Table 2: Number of lexems added to the lexicon.

# 4 Morphotactics

In this section, we introduce morphotactic rules of Standard Central Kurdish (SCK). The morphotactics are the rules that define the possible bound morphemes and their order in each word type. We have adopted the SCK morphotactic from the reliable linguistic works, especially from the elaborate descriptive study of Kurdish by MacKenzie (1961). Additionally, generative derivational affixes of CK are discussed. In discussing the morphological rules, we acquire a semantics-based perspective to words at the same time, where semantic-based limitations of CK morphotactics is taken into consideration. As an instance of semantic exceptions to the verb morphotactics of SCK, the verb */zanîn/* 'to know' is not considered well-formed when accepting direct objects other than the third person.

## 4.1 Nouns

The inflection rules of nouns in SCK is comprehensively studies and categorized, as can be seen in Table 3. Such inflections include suffixes and clitics since CK nouns do not accept any prefixes. Figure 1 shows the morphotactics of SCK nouns. All nodes can be final and all arcs are optional and can be skipped (by zero (epsilon) arc).

| No | Suffix | Function | Example |
|---|---|---|---|
| 1 | /-eke/ | definite | /*guł*-**eke**/ '**the** flower' |
| 2 | /-êk/ | indefinite | /*guł*-**êk**/ '**a** flower' |
| 3 | /-e/ | discontinuous demonstrative | /**ew** *guł*-**e**/ '**that** flower' |
| 4 | /-an/ | plural | /*guł*-**an**/ 'flower**s**' |
| 5 | /-îş/ | adverbial enclitic | /*guł*-**îş**/ 'flower **also**' |
| 6 | /-im\|-it\|-î\|-man\|-tan\|-yan/ | possessive pronoun | /*guł-eke*-**tan** *cwane*/ '**your** flower is beautiful' |
| 7 | /-im\|-it\|-î\|-man\|-tan\|-yan/ | argument pronoun (subject, object, or complement) | /*guł*-**man** *çinî*/ '**we** picked flowers', /*êwe guł*-**man** *pê deden*/ 'you give **us** flowers' |
| 8 | /-da/, /-ewe/ | circumposition | /**le** *guł*-**da**/ '**in** flower', /**le** *guł*-**ewe**/ '**from** flower' |
| 9 | /-î/ | izafe | /*guł*-**î** *Soran*/ 'flower **of** Soran' |
| 10 | /-e/ | izafe | /*guł*-**e** *zerd*eke/ '**the** yellow flower' |
| 11 | /-im\|-ît\|-e\|-în\|-in\|-in/ | copula | /*guł*-**în**/ '**we are** flowers' |
| 12 | /-e/, /-ê/ (only fem.), /-îne/ (plural) | vocative | /*ker*-**e**/ 'hey you ass!', /*kiç*-**ê**/ 'hey girl!', /*kiç*-**îne**/ 'hey girls!' |
| 13 | /-ê/ | locative | /*der*-**ê**/ 'in outside', /*şew*-**ê**/ 'at night' |

Table 3: The suffixes and clitics of SCK nouns.

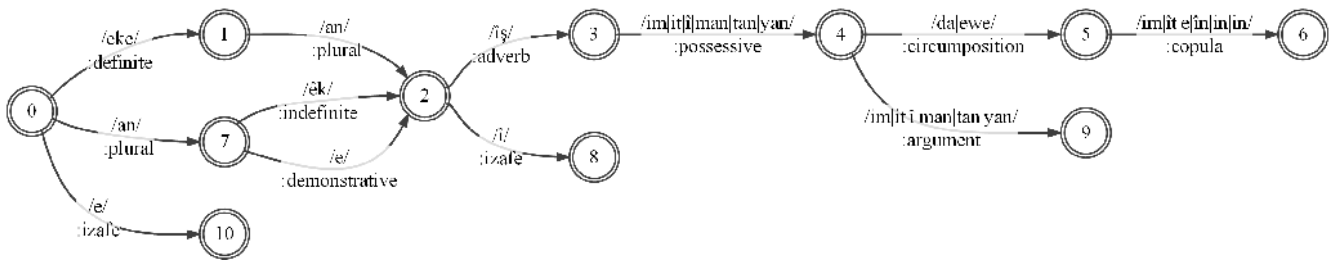

Figure 1: The morphotactics of general nouns in Standard Central Kurdish.

Notes on noun morphotactics:

- *Vocative and locative suffixes*: Vocative and locative case suffixes are not included in the description of general morphotactics in this paper. This is due to the fact that such suffixes are rare in formal text and are accepted only by a certain set of noun stems.

- *Dialect-specific suffixes*: Most dialects of CK do not have grammatical gender suffixes. Some northern dialects of CK, however, have suffixes for showing grammatical gender in oblique case or in izafe structure (MacKenzie, 1961, §179). Additionally, in northern dialects of CK as Arbili and Mukri, the suffix */-řa/* is the second part of some circumposition; e.g., */le hewlêr-řa/* 'from Arbil'. Since these suffixes do not appear in all dialects, especially in formal writings, we do not implement them.

- *Plural suffix position*: The suffixes of definite (*/-eke/*), indefinite (*/-êk/*) and discontinuous demonstrative (*/-e/*) are in complementary distribution. The plural suffix */an/* precedes */-êk/* and */-e/* (*/-anêk/* and */-ane/*) and proceeds */-eke/* (*/-ekan/*).

- *Plural suffix /-gel/*: Suffix */-gel/* (also */-el/* and */-eyl/*, in some dialects) is a plural suffix commonly used in some southern dialects of CK. In SCK, however, it only comes with the name of herds like */gagel/* 'herd of cows' or */mêgel/* 'herd of sheep' (MacKenzie, 1961, §177). This plural suffix */-gel/* is used rarely in SCK and results in a marked form, indicating the effect of the southern Kurdish dialects or the Persian syntax on the text. For example, */keştîgelî serbazî/* 'military ships' is a marked form and its well-formed and unmarked equivalent is */keştîye serbazîyekan/*.

- *Loaned plural suffixes*: The Arabic plural suffixes of */-at/*, */-hat/*, and */-cat/* as used in words */baẍat/*, */dêhat/*, and */mîwecat/* are only seen in calques (MacKenzie, 1961, §177) and usually do not have plural meaning. Therefore, we consider such forms as base forms since they can be suffixed with */-an/* suffix, as */dêhat-ekan/* 'villages'.

- *Clitic /-îş/*: In SCK, the clitic */-îş/* precedes the personal pronoun suffixes (MacKenzie, 1961, §240). However, this order is different in some southern dialects of CK. We take the SCK form as the well-formed.

- *Third person copula*: In rare cases, the third-person singular copula (*/-e/*) may occur after subject or object personal pronoun suffixes, in which case it means 'to want'; e.g., */bo guł-eke-m-tan-e?/* has the same meaning as the standard from */bo guł-eke-m-tan dewêt?/* 'Do you want it for my flower?'. Furthermore, if third person singular copula (*/-e/*) occurs immediately after third person personal pronoun suffix (*/î/*), they swap with an intervening consonant */-t-/* between them; e.g., */mał-yan-e/* 'it is their home', but */mał-ye-tî/* 'it is his home'.

- */-ewe/ with location words*: The derivational suffix */-ewe/* in some location words like */jûrewe/* 'inside' and */małewe/* 'inside' shouldn't be confused with the inflectional suffix */-ewe/* (Edmonds, 1955).

### 4.1.1 Named Entities, Titles, and Units of Measurement

The proper noun or so-called named entities can accept all the suffixes in Table 3. When a named entity is used as a general noun, it can take definite, indefinite, discontinuous demonstrative, and plural suffixes; e.g., */hewlêr-eke-m/* 'my Erbil'. Although there is no difference between general and proper nouns in terms of inflection, we tagged named entities by five categories in this study, for future applications of the analyzer (such as automatic data extraction):

- Location: */kurdistan/* 'Kurdistan', */hewlêr/* 'Erbil'.
- Person: */ʔaso/* 'Aso', */barzanî/* 'Barzani'.
- Organization: */pêşmerge/* 'Peshmerga', */fîfa/* 'FIFA'.
- Time/Date: */çwarşemme/* 'Wednesday', */ʔab/* 'August'.
- Miscellaneous: */dolar/* 'dollar', */qurʔan/* 'Quran'.

If a noun is used as a title, it cannot accept affixes. There are two types of titles based on appearance in utterance: A) Immediately before a proper name; e.g., */kak/* 'Mr.', */şêx/* 'Sheikh'. B) Immediately after a proper name; e.g., */xanim/* 'Ms.', */beg/* 'Bey'. Units of measurement usually occur uninflected between a number and a noun; e.g., */dû kîlo goşt/* 'two kilograms of meat'. Units can also have suffixes; e.g., */bist-êk-îş-e/* 'it is also a span'. The morphotactics of the units of measurement is similar to nouns except they do not accept plural suffixes.

## 4.2 Adjectives

There are several categories of adjectives in SCK. Here, such categories are illustrated with an example. Then, the morphotactics of each is elaborated on. The categories of SCK are as follows:

- Descriptive: e.g., */sûr/* 'red', */gewre/* 'big', */behêz/* 'strong', */řîşsipî/* 'gray bearded'
- Quantitative: e.g., */yek/* 'one', */dû/* 'two', */sed/* 'hundred', */hîç/* 'none', */çend/* 'several'
- Indefinite: e.g., */zor/* 'many/much', */kem/* 'few/little', */hendêk/* 'some', */çend/* 'several'
- Demonstrative: e.g., */ʔem ...-e/* 'this', */ʔew ...-e/* 'that'
- Interrogative: e.g., */kam/* 'which', */çi/* 'what'
- Distributive: e.g., */her/* 'every', */hîç/* 'none'

Demonstrative, interrogative and distributive adjectives are indeclinable, i.e., they do not accept any inflection. In the following, we elaborate on the morphotactics of descriptive and quantitative adjective which are similar to nouns' morphotactics. Also, some exceptional adjectives that only take special affixes are described.

### 4.2.1 Descriptive Adjectives

As descriptive adjectives can be converted to nouns by zero derivation, the suffixes they accept as well as their order are the same as nouns. Additionally, gradable descriptive adjectives can take suffixes */-tir/* and */-tirîn/* for comparative and superlative forms respectively, contiguous to the stem. In order to prevent ill-formed comparative and superlative forms involving non-gradable adjectives, like */\*win-tir/*, we have tagged the non-gradable descriptive adjectives in our lexicon, such as */gyandar/* 'living' and */win/* 'missing'. It is noteworthy that the indefinite suffix */-êk/* means that an adjective has been converted to a noun.

Descriptive adjectives in SCK are used in the following structures: 1) /guł-î zerd/ 'yellow flower'; 2) /guł-e zerd-eke/ 'the yellow flower'; and, 3) less commonly /zerd-e guł/ 'yellow flower' (Karimi, 2007).

Comparative adjectives come after nouns and accept all noun suffixes and clitics; e.g., /çya berz-tir-ekan/ 'the higher mountains'. Superlative adjectives also come before the nouns but do not accept any affixes or clitics; however, they can be converted to nouns with zero derivation and in such cases, they have similar inflection to nouns; e.g., /nizîk-tirîn-eke-yan-în/ 'we are their nearest one'. There is another way to create superlative adjectives by using /here/ 'the most' before the descriptive adjectives (usually with definite suffix); e.g., /kuř-e here berz-eke-yan/ 'their tallest boy'.

Furthermore, we have recorded the following types of proper descriptive adjectives with different tags, for future named entity recognition tasks: A) Languages and dialects; e.g., /kurdî/ 'Kurdish', /badînî/ 'Badini'. B) Ethnic groups; e.g., /kurd/ 'Kurd', /ʔermen/ 'Armenian'. C) Demonyms; e.g., /ʔemrîkî/ 'American', /sûrî/ 'Syrian'.

### 4.2.2 Quantitative Adjectives

The numerals, like descriptive adjectives, can be converted to nouns by zero derivation; e.g., /şeş-êk-î tir/ 'another six'. The ordinal numbers are formed by adding suffixes /-em/ and /-emîn/ to the cardinal numbers (MacKenzie, 1961, §195); e.g., /şeş car/ meaning 'six times', and /şeş-em car/, /şeş-emîn car/ and /carî şeş-em/ all meaning 'the sixth time'. The ordinal number preceding the modified noun is indeclinable; however, if it comes after the nouns, its inflection rules match the descriptive adjectives; e.g., /carî şeş-em-eke-ş-tan/ 'also your sixth time'. The powers of 10 can get plural suffix /an/, as in /de-yan/ 'tens' and /sed-an/ 'hundreds', and can be used as a noun; e.g., /sed-an kes/ 'hundreds of people'.

### 4.2.3 Exceptional Adjectives

There is a set of adjectives that are exceptional in the sense that do not accept /-eke/, /-êk/, /-an/, /-e/ (demonstrative) and /-e/ (izafe) suffixes; e.g., /le bîrî kesêkî tir-îş-da-yn/ 'we are also thinking of someone else', /kesêkî weha-ş-man nîye/ 'we do not have such a person'. These adjectives are /tir/, /dîke/, /ke/ and /dî/ meaning 'other', 'another' and 'else'; as well as /weha/ and /wa/ meaning 'such'. Additionally, the adjective /weha/ also precedes nouns but cannot be inflected; e.g., /weha kesêk/ 'such a person'.

## 4.3 Pronouns

The pronouns in CK can be categorized as follows:

- Personal: e.g., /min/ 'I', /ʔewan/ 'they'.
- Demonstrative: /ʔem/ 'this' and /ʔew/ 'that'.
- Reflexive: /xo-/ '-self'.
- Reciprocal: /yektir/ 'each other'.
- Interrogative: e.g., /çi/çî/ 'what', /çon/ 'how', /çend/ 'how many', /kê/ 'who', /key/ 'when', /kam(e)/ 'which', and /kwê/ 'where'. /kwa/ 'where is it?' contains the sense of a verb (MacKenzie, 1961, §203).
- Indefinite: e.g., /hendêk/ 'some', /hîç/ 'none', /gişt/ 'all'.

In what follows, we draw upon members of each of those categories and elaborate on their inflectional rules as well as exceptions.

### 4.3.1 Personal Pronouns

In CK, there are three persons and two numbers. Therefore, a total of six person-number pronouns exist in the language. Table 4 shows the different sets of pronouns and agreement markers and general function of each set. The agreements and pronoun's functions may change in some ditransitive past tense verbs (Haig, 2004). Figure 2 shows the morphotactics of separate personal pronouns.

As mentioned earlier, there is no gender in standard CK. Additionally, there is another structure, i.e., pronominal clitic, that happens only in some dialects of CK, which we will not focus on. This structure occurs in southern dialects of CK such as Sanandaji and Jaffi and indicates the object of transitive past verbs. As an instance, the form /nard-man-tan/ 'you sent us' is equal to standard form of /nard-tan-în/.

### 4.3.2 Demonstrative Pronouns

Corresponding to the demonstrative adjectives /ʔem ...-e/ 'this', /ʔew ...-e/ 'that', there are two demonstrative pronouns in SCK: /ʔem/ 'this' and /ʔew/ 'that' (MacKenzie, 1961, §202). Figure 3 shows the morphotactics of demonstrative pronouns. Due to very rare occurrence, the indefinite suffix for demonstrative pronouns (e.g., /bo ʔem-êk ke daykî nebînîwe/ 'for one this who has not seen his/her/its mother') has not been implemented.

The indeclinable pronoun /hî/ 'that of (possessed by)' appears before nouns, pronouns and adjectives and forms a demonstrative izafe structure (MacKenzie, 1961, §188); e.g., /hî birakem/ 'that of my brother', /hî ʔême/ 'ours'.

| Set | 1SG | 2SG | 3SG | 1PL | 2PL | 3PL | Functions |
|---|---|---|---|---|---|---|---|
| 1 | /min/ | /to/ | /ʔew/ | /ʔême/ | /ʔêwe/ | /ʔewan/ | -Separate Subject/Object |
| 2 | /-im/ | /-it/ | /-î/ | /-man/ | /-tan/ | /-yan/ | pronominal clitics:<br>- Possessive<br>- Subject of transitive past verbs<br>- Object of present and imperative verbs |
| 3 | /-im/ | /-ît/ | /-êt/ | /-în/ | /-in/ | /-in/ | -Subject of present verbs |
| 4 | /-im/ | /-ît/ | Ø | /-în/ | /-in/ | /-in/ | -Subject of intransitive past verbs<br>- Object of transitive past verbs |
| 5 | /-im/ | /-ît/ | /-e/ | /-în/ | /-in/ | /-in/ | - Object of transitive present perfect verbs<br>- Copula |
| 6 | - | /e/ | - | - | /-in/ | - | - Subject of imperative verbs |

Table 4: Different sets of pronouns and agreement markers in SCK

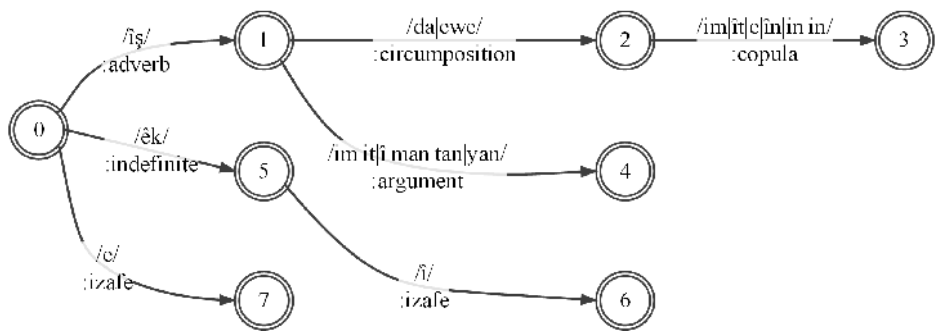

Figure 2: The morphotactics of separate personal pronouns

### 4.3.3 Other Pronouns

The morphotactics of reflexive, reciprocal, interrogative and indefinite pronouns generally follows the structure shown in Figure 4. However, the reciprocal pronoun */yektir/* 'each other' and interrogative pronoun */key/* 'when' do not accept possessive suffixes. Also, uninflected reflexive pronoun */xo-/* 'self, can be changed to */xud/*, which with izafe clitic */î/* becomes */xud-î/*.

## 4.4 Verbs

In this section, we examine the rules of SCK verb inflection and their exceptions. In CK morphology, verbs have complicated structures as they accept various affixes and clitics. Additionally, ergativity has made the structure of CK verbs more complex than neighboring languages such as Persian and Turkish []. In short, the ergativity in SCK makes the past tense pronoun suffixes different in the intransitive and transitive verbs (Kareem, 2016) (Karimi, 2007). Thus, along with learning the stems of each verb, one must also remember its transitivity. Furthermore, another reason for complexity of SCK verb inflections is the existence of exceptional verbs that do not accept direct objects. These verbs include both intransitive verbs, such as onomatopoeic verbs, and the transitive verbs, like */řwanîn/* 'to look' and */koşîn/* 'to try'.

### 4.4.1 Verb Stems

Works that have studied CK morphology have considered two stems of past and present for each verb, based on which tense they are used for. Although denoted past or present, these stems do not carry the tense-specific characteristics. Some linguists (Anoushe, 2018) (Baban, 2012) believe that the past stem is made by adding past suffix to the present stem. However, due to lack of the real meaning of time in the stems (Kareem, 2016) and the existence of numerous exceptions, we in the current study consider the present verb stems as irregular, hence creating a lookup table for present stems. Also, we follow the existing literature in using the terms 'past' and 'present' for verb stems.

In a few cases, certain verbs have multiple past or present stems as a result of suppletion or phonological changes. Table 5 lists such verbs with their multiple past or present stems.

In some intransitive verbs, the final vowel of the past stem varies between */a/* and */î/*; e.g., */qewman/* -/qewmîn/ 'to happen', */řiman/-/řimîn/* 'to collapse', */leran-ewe/-/lerîn-ewe/* 'to shake'.

Due to ergativity of CK, verb inflection is affected by transitivity, that is, the pronoun suffixes are different for intransitive and transitive past verbs; therefore, it is necessary to determine the transitivity of each verb in the lexicon.

### 4.4.2 Causative Verbs

In CK, the causative verb is made from the intransitive present stem. The past stem of the causative verb is made by adding */-and/* and the present stem is made by adding */-ên/* to the causative stem. For example, for intransitive verb */xewtin/* 'to sleep', the causative stem is same as the present stem /xew/. Therefore, the causative past stem is /xew-and/ and the causative present stem is /xew-ên/. There are some intransitive verbs, however, that cannot be made causative, since they have an independent transitive equivalent; take the following pairs as an example: /kewtin/ 'to fall' and /xistin/ 'to throw/to make fall', /řoyştin/ 'to go' and /birdin/ 'to carry', /hatin/ 'to come' and /hênain/ 'to bring', /man/ 'to stay' and /hêştin/ 'to hold. We record the possible causative stem for each intransitive verb in our lexicon.

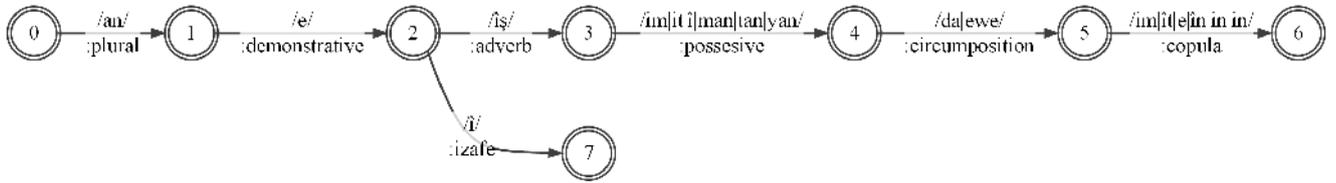

Figure 3: The morphotactics of demonstrative pronouns /ʔem/ and /ʔew/

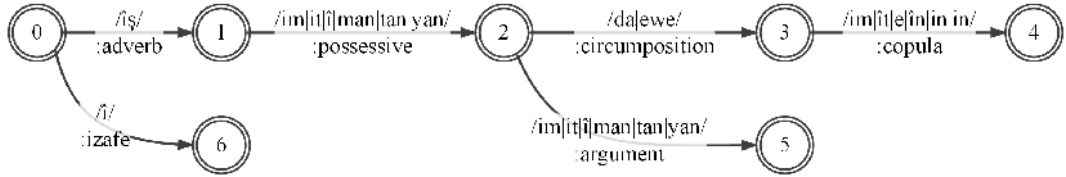

Figure 4: The morphotactics of reflexive, reciprocal, interrogative, and indefinite pronouns.

There are some morphophonological rules in formation of causative stems from the present stems as follow:

- *Stems ending in /ê/*: for intransitive present stems ending in /ê/, as in /qewma:qewmê/ 'to happen', the final vowel /ê/ is dropped in the causative stem, as /qewm-and:qewm-ên/ 'to make happen'.

- *Stems ending in /e/*: for intransitive present stems ending in /e/, as in /geyşt:ge/ 'to reach', an epenthesis of /y/ occurs in their causative stems, as /gey-and:gey-ên/ 'to convey'.

- *Stems ending in /î/, /u/ or /û/*: for intransitive present stems ending in vowels /î/, /u/ or /û/, as in /jya:jî/ 'to live' and /nûst:nû/ 'to sleep', the final vowel turns into equivalent approximant (/y/ or /w/) in their causative stems, as /jy-and:jy-ên/ 'to keep alive' and /nw-and:nw-ên/ 'to put to sleep'.

### 4.4.3 Passive Verbs

In CK, all transitive verbs have a passive voice which is inflectional and is conjugated similar to other intransitive verbs. Passive stems are mainly made from the present stem of the transitive verbs and suffix /-r-/. For example, the present stem of the verb /kuştin/ 'to kill' is /kuj-/, and its passive stem is /kujr-/. There are some exceptions to this rule as mentioned later in this section. The passive past stem is made by adding /-a/ (as /kujr-a-m/ 'I was killed') and passive present stem is made through adding /-ê/ (as /de-kujr-ê-m/ 'I will be killed') to the passive stem. The passive verbs are also made from causative stems. For example, the causative verb /xew-and-in/ is changed to form its passive tense /xew-ên-r-a-n/ 'to be put to sleep'. It's noteworthy that the causative verb /xew-and-in/ itself is formed from intransitive verb /xewt-in/ 'to sleep'.

Due to the complex morphophonological rules and numerous exceptions in the production of the passive stem, they were recorded in our lexicon. The passive past and present stems are, however, automatically generated by adding /-a/ and /-ê/ respectively. There are some morphophonological rules and irregularities in formation of passive stems are as follow:

- *Active present stem ending in /e/, /o/ or /ê/*: in such cases, the ending vowel becomes /i/, which does not appear in the written form: e.g., /dir/</de-r/ 'to be given', /bir/</be-r/ 'to be taken', /kir/</ke-r/ 'to be done', /xir/</xe-r/ 'to be thrown', /xur/</xo-r/ 'to be eaten', /nir/</ne-r/ 'to be set'.

- *Active present stem ending in consonants /r/ and /ř/*: in such cases, the passive suffix /r/ is dropped; e.g., /biř/</biř-ra/ 'to be cut', /bijêr/</bijêr-r/ 'to be selected'.

Irregular exceptions happen in formation of passive for the following verbs:

- Active present stem in verb /girtin/ 'to take' is /gir/, but the passive stem is /gîr/.
- Active present stem in verb /gutin/ 'to say' is /łê/ but the passive stem is /gutr/.
- Active present stem in verb /wîstin/ 'to ask' is /ewê/, but the passive stem is /wîstr/.

### 4.4.4 Onomatopoeic Verbs

Onomatopoeic verbs in CK are made up of an onomatopoeic stem, such as /cîk-/ (the sound of sparrow) and their past and present stems, similar to the causative verbs, is made by adding /-and/ and /-ên/, respectively, resulting in /cîkand/ and /cîkên/. This structure is generative and can be used to construct a verb with almost any sound. These verbs are semantically intransitive and do not take objects, but are conjugated in the past tense, similar to transitive verbs (MacKenzie, 1961).

| Verb | Past Stems | Present Stems |
|---|---|---|
| 'to hear' | /bîst/ | /biye/, /bîst/ (/bîs/) |
| 'to carry' | /bird/ | /be/, /ber/ |
| 'to see' | /dît/ (/dî/), /bînî/ | /bîn/ |
| 'to arrive' | /geyşt/, /geyî/ | /ge/ |
| 'to go' | /řoyşt/, /řoyî/ | /řo/ |
| 'to bring' | /hêna/, /hanî/ | /hên/ |
| 'to send' | /nard/, /henard/ | /nêr/ |
| 'to bite' | /gest/, /gezî/ | /gez/ |
| 'to give birth' | /za/, /zayî/ | /zê/ |
| 'to wash' | /şit/, /şord/ | /şo/, /şor/ |
| 'to transport' | /gwast/, /gwêza/ | /gwaz/, /gwêz/ |
| 'to snatch' | /řifand/, /fiřand/ | /řifên/, /fiřên/ |

Table 5: Verbs with multiple past and/or present stems.

Therefore, we categorize them in a group separate from transitive and intransitive verbs. The onomatopoeic verbs do not have passive or causative equivalents and do not accept preverbs.

### 4.4.5 Verbal Affixes and Clitics

The affixes and clitics of CK verbs are:

- *Preverb*: preverbs are bound adverbs that change the meaning of the verb. The most frequent preverbs are: */heł-/*, */da-/*, */řa-/*, */der-/*, */řo-/*, and */wer-/*. Some verbs come only with preverbs, for instance, the verb */\*wasîn/* never comes separately; but with preverb */heł-/*, it is changed to */heł-wasîn/* which means 'to hang up, to suspend'.

- *Postverb /-ewe/*: all simple verbs can take */-ewe/* to convey the meaning of repetition or return, but some verbs, such as */şardin-ewe/* 'to hide', are meaningless without */-ewe/* (*/\*şardin/*). For those verbs, the compulsory */-ewe/* must be indicated in the lexicon. The suffix */-ewe/* gives one of the following meanings to the verb (MacKenzie, 1961, §235):

  - *Repetition*; e.g., */deygirim/* 'I get it', */deygirim-ewe/* 'I get it again'
  - *Return*; e.g., */dam/* 'I gave' */dam-ewe/* 'I gave back'
  - *Different new meaning*; e.g., */xwardim/* 'I ate' */xwardim-ewe/* 'I drank'

- *Postverb /-e/*: postverb */-e/* can replace prepositions */be/* or */bo/* if a verb has the valency of a prepositional phrase. In this case, it is compulsory for a noun phrase to come after the verb (MacKenzie, 1961, §236); e.g., */deçim bo derewe/* 'I go outside' becomes */deçim-e derewe/* 'I go outside'.

- *Negation and tense-mood-aspect affixes*: in verb structure of CK, defining just one function for each part is not always possible. For example, depending on the tense-aspect of the verbs, prefixes of */ne-/*, */na-/* or */me-/* show the negation. However, in addition to negating the verbs, they may show the mood (as */ne-xom/* 'if I do not eat') and aspect (as */na-xom/* 'I do not eat'). Following affixes are used in verb conjugation to form negation, tense, mood, or aspect of the verbs:

  - The prefix */de-/* denotes the imperfective aspect.
  - The prefix */bi-/* is used in some tenses to form subjunctive mood.
  - The suffix */-ib/* and */ibêt/* are used in some tenses to form subjunctive mood.
  - The suffix */-û/* is used in present perfect aspect.
  - The suffix */-ibû/* is used in past perfect aspect.
  - The suffix */-aye/* is used in subjunctive imperfect and subjunctive present perfect tenses.

- *Adverbial clitic /-îş/*: add the meaning of 'also, too, even'.
- *Pronominal clitics /im|it|î|man|tan|yan/*: indicating the object in present tenses and subject in past tenses of transitive verbs.

If the clitics */-îş/* and/or */im|it|î|man|tan|yan/* are present in the sentence, they must be connected to the first existing in the sentence among: separated object, preposition, preverb, negative/subjunctive prefix, and imperfect prefix (McCarus, 1958, pp. 103–106). If both clitics exist, */-îş/* immediately precedes */im|it|î|man|tan|yan/* (MacKenzie, 1961, §240).

### 4.4.6 Present and Imperative Verbs

The present and imperative verbs are made up from present stem. Table 6 shows the inflection structure of present and imperative verbs. Note that in the following tables, constituents without parenthesis are compulsory; i.e., parenthesis show optionality.

| Structure | -2 | -1 | 0 | +1 | Examples |
|---|---|---|---|---|---|
| present (indicative) | /de-\|na-/ | (clitic) | stem | /-im\|-ît\|-êt\|-în\|-in/ | /de-ş-kew-êt/ /de-ş-man-nêr-êt/ |
| present (subjunctive) | /bi-\|ne-/ | (clitic) | stem | /-im\|-ît\|-êt\|-în\|-in/ | /bi-ş-kew-êt/ /bi-ş-man-nêr-êt/ |
| imperative | /bi-\|me-/ | (clitic) | stem | /-e\|-in/ | /bi-ş-kew-e/ /bi-ş-man-nêr-e/ |

Table 6: Order of components in present and imperative verbs. The preverb in position -3 and the postverb in position +2.

The optional clitics (position "-1") can be filled by /îş/ and in transitive verbs by the object pronominal clitic (/im|it|î|man|tan|yan/). These two clitics are displaced according to the syntax and they attach to the leftmost constituent of their phrases (Haig, 2004). Table 7 shows the placement of these clitics in utterances with a transitive present verb.

| Example | Sep. Obj. | Prep. | Preverb | Neg./Mood/Aspect | Stem | Subject |
|---|---|---|---|---|---|---|
| /de-ş-tan-gir-în/ | | | | /de-@/ | /gir/ | /-în/ |
| /heł-îş-tan de-gir-în/ | | | /heł-@/ | /de-/ | /gir/ | /-în/ |
| /bo-ş-tan heł de-gir-în/ | | /bo-@/ | /heł-/ | /de-/ | /gir/ | /-în/ |
| /cil-îş-tan bo heł de-gir-în/ | /cil-@/ | /bo/ | /heł-/ | /de-/ | /gir/ | /-în/ |

Table 7: placement of clitics /-îş/ and /im|it|î|man|tan|yan/ in utterances with a transitive present verb.

If the verb has a preverb or is compound, the /bi-/ prefix is frequently dropped in positive imperative and subjunctive present tenses; e.g., /da-nîş-in/ < /da-bi-nîş-in/ 'sit down!', /ba gwê-y bo gir-în/ < /bi-gir-în/ 'let's listen to it'

There are two irregularities with present and imperative verbs in CK. The positive imperative form of the verb /hatin/ 'to come' is irregular, as the present stem is /-ê-/, but imperative forms are /were/ '(singular) come!' and /werin/ '(plural) come!'. Another irregularity occurs in the present paradigms of the verbs /wîstin/ 'to want' (with present stem /ewê/). Unlike the other verbs, in this verb, clitics /im|it|î|man|tan|yan/ indicate the subject and pronouns /im|ît|Ø|în|in|in/ indicate the object; e.g., /de-man-ewê-n/ 'we want them'. Table 8 shows the morphotactic of present tenses of verb /wîstin/ 'to want'.

| Tense/Aspect | Mood | -2 | -1 | 0 | +1 | Example |
|---|---|---|---|---|---|---|
| simple | Ind. | /de-\|na-/ | (clitic) | /ewê/ | /-im\|-ît\|-êt\|-în\|-in/ | /de-ş-man-ewê-n/ |
| simple | Subj. | /bi-\|ne-/ | (clitic) | /ewê/ | /-im\|-ît\|-êt\|-în\|-in/ | /bi-ş-man-ewê-n/ |

Table 8: Inflection of present verbs of /wîstin/ 'to want'. The postverb occur in position +2.

### 4.4.7 Past Tense Verbs

Due to the ergativity of CK (Kareem, 2016), verbal inflection in past tenses is affected by transitivity, that is, the pronouns are different for past tenses of transitive verbs:

- Intransitive present: /de-kew-**în**/ 'we fall'
- Intransitive past: /de-kewt-**în**/ 'we were falling'
- Transitive present: /de-yan-nêr-**în**/ 'we send them'
- Transitive past: /de-**man**-nard-in/ 'we were sending them'

That is, the pronoun set of /im|ît|Ø|în|in|in/ indicates the subject and clitic set of /im|it|î|man|tan|yan/ indicates the object. However, the functions switch for transitive past and the verb agrees with the object (Karimi, 2010). In addition to transitive verbs, onomatopoeic verbs such as /qîřandin/ 'to scream' and /nałandin/ 'to groan' are conjugated like transitive verbs; e.g., /qîřand-man/ 'we screamed' and /de-man-qîřand/ 'we were screaming'.

There are some verbs such as /řwanîn/ (/nwařîn/) 'to look', /wêran/ 'to dare', and /koşîn/ (/koşan/) 'to strive', which are transitive but only accept indirect prepositional object. For example, /*de-man-řwanî-n/ is ill-formed, but /de-man-řwanî bo ewan/ 'we were looking at them' is well-formed.

We have summarized the general structure of the CK past verb in the following tables. Table 9 shows the order of component in all tenses of intransitive past verbs. Table 10 shows the order of component in all tenses of transitive past verbs. Note that there is no subjunctive mood for simple past of transitive verbs.

In northern dialects of CK, as Mukri, the structure of subjunctive imperfect and subjunctive past perfect are different. They are made by suffix /-ba/ in position +1 and removing /-aye/ in position +3. For example, the intransitive past verbs /bi-kewt-iba-yn/ and /kewt-ibû-ba-yn/ are equivalent to /bi-kewt-în-aye/ and /kewt-ibû-yn-aye/, respectively; and the transitive past verbs /bi-man-xist-iba-yt/ and /xist-ibû-ba-man-ît/ are equivalent to /bi-man-xist-ît-aye/ and /xist-ibû-man-ît-aye/, respectively.

As mentioned earlier, clitics /-îş/ and /im|it|î|man|tan|yan/ are displaced according to the syntax. Table 11 shows the placement of these clitics in utterances with a transitive past verb.

| Tense/Aspect | Polarity | Mood | -2 | -1 | 0 | +1 | +2 | +3 | Example |
|---|---|---|---|---|---|---|---|---|---|
| simple | P | Ind. | - | - | stem | - | /-im/-ît/Ø/-în/-in/ | - | /kewt-în/ |
| simple | N | Ind. | /ne-/ | - | stem | - | /-im/-ît/Ø/-în/-in/ | - | /ne-kewt-în/ |
| simple | P | Subj. | - | - | stem | /-ib-/ | /-im/-ît/Ø/-în/-in/ | - | /kewt-ib-în/ |
| simple | N | Subj. | /ne-/ | - | stem | /-ib-/ | /-im/-ît/Ø/-în/-in/ | - | /ne-kewt-ib-în/ |
| imperfect | P | Ind. | - | /de-/ | stem | - | /-im/-ît/Ø/-în/-in/ | - | /de-kewt-în/ |
| imperfect | N | Ind. | /ne-/ | /de-/ | stem | - | /-im/-ît/Ø/-în/-in/ | - | /ne-de-kewt-în/ |
| imperfect | P | Subj. | /bi-/ | - | stem | - | /-im/-ît/Ø/-în/-in/ | /-aye/ | /bi-kewt-în-aye/ |
| imperfect | N | Subj. | /ne-/ | - | stem | - | /-im/-ît/Ø/-în/-in/ | /-aye/ | /ne-kewt-în-aye/ |
| pr. perfect | P | Ind. | - | - | stem | /-û-/ | /-im/-ît/-e/-în/-in/ | - | /kewt-û-yn/ |
| pr. perfect | N | Ind. | /ne-/ | - | stem | /-û-/ | /-im/-ît/-e/-în/-in/ | - | /ne-kewt-û-yn/ |
| pr. perfect | P | Subj. | - | - | stem | /-ibêt-/ | /-im/-ît/Ø/-în/-in/ | - | /kewt-ibêt-în/ |
| pr. perfect | N | Subj. | /ne-/ | - | stem | /-ibêt-/ | /-im/-ît/Ø/-în/-in/ | - | /ne-kewt-bêt-în/ |
| ps. perfect | P | Ind. | - | - | stem | /-ibû-/ | /-im/-ît/Ø/-în/-in/ | - | /kewt-ibû-yn/ |
| ps. perfect | N | Ind. | /ne-/ | - | stem | /-ibû-/ | /-im/-ît/Ø/-în/-in/ | - | /ne-kewt-ibû-yn/ |
| ps. perfect | P | Subj. | - | - | stem | /-ibû-/ | /-im/-ît/Ø/-în/-in/ | /-aye/ | /kewt-ibû-yn-aye/ |
| ps. perfect | N | Subj. | /ne-/ | - | stem | /-ibû-/ | /-im/-ît/Ø/-în/-in/ | /-aye/ | /ne-kewt-ibû-yn-aye/ |

Table 9: Morphotactics of intransitive past verbs. The preverb in position -3 and the postverb in position +4

| Tense/Aspect | Polarity | Mood | -2 | -1 | 0 | +1 | +2 | +3 | Example |
|---|---|---|---|---|---|---|---|---|---|
| simple | P | Ind. | - | - | stem | - | /-im/-ît/Ø/-în/-in/ | - | /xist-man-ît/ |
| simple | N | Ind. | /ne-/ | - | stem | - | /-im/-ît/Ø/-în/-in/ | - | /ne-man-xist-ît/ |
| imperfect | P | Ind. | - | /de-/ | stem | - | /-im/-ît/Ø/-în/-in/ | - | /de-man-xist-ît/ |
| imperfect | N | Ind. | /ne-/ | /de-/ | stem | - | /-im/-ît/Ø/-în/-in/ | - | /ne-man-de-xist-ît/ |
| imperfect | P | Subj. | /bi-/ | - | stem | - | /-im/-ît/Ø/-în/-in/ | /-aye/ | /bi-man-xist-ît-aye/ |
| imperfect | N | Subj. | /ne-/ | - | stem | - | /-im/-ît/Ø/-în/-in/ | /-aye/ | /ne-man-xist-ît-aye/ |
| pr. perfect | P | Ind. | - | - | stem | /-û-/ | /-im/-ît/-e/-în/-in/ | - | /xist-û-man-ît/ |
| pr. perfect | N | Ind. | /ne-/ | - | stem | /-û-/ | /-im/-ît/-e/-în/-in/ | - | /ne-man-xist-û-yt/ |
| pr. perfect | P | Subj. | - | - | stem | /-ibêt-/ | /-im/-ît/Ø/-în/-in/ | - | /xist-ibêt-man-ît/ |
| pr. perfect | N | Subj. | /ne-/ | - | stem | /-ibêt-/ | /-im/-ît/Ø/-în/-in/ | - | /ne-man-xist-ibêt-ît/ |
| ps. perfect | P | Ind. | - | - | stem | /-ibû-/ | /-im/-ît/Ø/-în/-in/ | - | /xist-ibû-man-ît/ |
| ps. perfect | N | Ind. | /ne-/ | - | stem | /-ibû-/ | /-im/-ît/Ø/-în/-in/ | - | /ne-man-xist-ibû-yt/ |
| ps. perfect | P | Subj. | - | - | stem | /-ibû-/ | /-im/-ît/Ø/-în/-in/ | /-aye/ | /xist-ibû-man-ît-aye/ |
| ps. perfect | N | Subj. | /ne-/ | - | stem | /-ibû-/ | /-im/-ît/Ø/-în/-in/ | /-aye/ | /ne-man-xist-ibû-ît-aye/ |

Table 10: Morphotactics of transitive past verbs. The preverb in position -3 and the postverb in position +4.

#### 4.4.8 Special Verbs

Verbs in CK are generally inflected following regular sets of rules. There are some exceptions, however, that are mentioned in the following. The verb */hebûn/* 'to be, to exist' or 'to have' is an instance of such verbs. This verb with the meaning 'to be, to exist' is the combination of the bound morpheme */he-/* and the verb */bûn/* and is mostly used in the third person singular form (MacKenzie, 1961, §218). The separated form of indicative present tense of the verb */hebûn/*, i.e., copula verb, occurs only in emphatic cases. For example, when there are no emphasis, */berz heyn/* 'we are tall' is ill-formed, and we should say */berz-în/*. Table 12 shows the order of component in the verbs */hebûn/* 'to be, to exist'.

The verb */hebûn/* with meaning 'to have' is transitive and uses clitics */im|it|î|man|tan|yan/* to indicate agent. The pronoun set of */im|ît|Ø|în|in|in/* are rarely used to indicate the object; e.g., */he-man-in/* 'we have them'. Table 13 shows the order of component in the verbs */hebûn/* 'to have'.

As mentioned for transitive past tenses, whenever the third singular subject pronouns occur adjacent to the object pronoun, their position is swapped. The paradigms of indicative present tense of */hebûn/* with third singular subject are: */he-m-î/*, */he-yt-î/*, */he-ye-tî/*, */he-yn-î/* and */he-n-î/*. In */he-ye-tî/* 'he has it' two epentheses occur: */-y-/* between two */e/*s and */-t-/* between */e/* and */î/* (Baban, 2012). A similar structure is also used in some words such as */birsî/* 'hungry' and */germa/* 'heat', and */bes/* 'enough' (MacKenzie, 1961, §209 and §219). For example, */birsî-ye/* means 'he/she/it is hungry', but */birsî-ye-tî/* means than 'he/she/it has the quality of being hungry'. Table 14 shows the difference between "to be" an adjective and "to have" a quality.

The noun */birîtî/* is only used in the frequent structures of */birîtî-ye le/* 'is consisted of' and */birîtî-n le/* 'are consisted of' and the compound verb of */birîtî-bûn/* 'to be consisted of'.

### 4.5 Prepositions

In terms of inflection, we divide the prepositions of CK into the following groups:

- Indeclinable: */be/* 'to', */le/* 'at, from', */(be)bê/* 'without', */wek(û)/* 'like', */berew/* 'toward', */(he)ta(kû)/* 'until', and */lemeř/* 'about'.

- Accepting argument pronoun clitic: */pê/* 'to', */lê/* 'at, from', */tê/* 'inside', */bo/* 'for'.

- Not accepting argument clitic: */legeł/* 'with', */pêş/* 'before', */paş/* 'after', */nêwan/* 'between', */ber/* 'front', */bin/* 'beneath',

- Similar to previous group but not occurring separately: */lebare-/* 'about', */bedwa-/* 'after', */ledwa-/* 'around'.

| Example | Sep. Object | Prep. | Preverb | Neg. | Imperfect | Stem | Aux. | Object |
|---|---|---|---|---|---|---|---|---|
| /xist-ibû-ş-tan-în/ | | | | | | /xist/ | /-ibû-@/ | /-în/ |
| /de-ş-tan-xist-în/ | | | | | /de-@/ | /xist/ | | /-în/ |
| /ne-ş-tan-de-xist-în/ | | | | /ne-@/ | /de-/ | /xist/ | | /-în/ |
| /heł-îş-tan ne-de-xist-în/ | | | /heł-@/ | /ne-/ | /de-/ | /xist/ | | /-în/ |
| /bo-ş-tan heł ne-de-xist-în/ | | /bo-@/ | /heł-/ | /ne-/ | /de-/ | /xist/ | | /-în/ |
| /cil-îş-tan bo heł ne-de-xist-în/ | /cil-@/ | /bo/ | /heł-/ | /ne-/ | /de-/ | /xist/ | | /-în/ |

Table 11: Placement of clitics /-îş/ and /im|it|î|man|tan|yan/ in some utterances with a transitive past verb.

| Tense/Aspect | Mood | 0 | +1 | +2 | +3 | Example |
|---|---|---|---|---|---|---|
| present | Ind. | /he-|nî-/ | - | /-im|-ît|-e|-în|-in/ | - | /heyn/ |
| present | Subj. | /he-|ne-/ | /-ib/ | /-im|-ît|-êt|-în|-in/ | - | /hebîn/ |
| simple past | Ind. | /he-|ne-/ | /-ibû-/ | /-im|-ît|Ø|-în|-in/ | - | /hebûyn/ |
| simple past | Subj. | /he-|ne-/ | /-ibû-/ | /-im|-ît|Ø|-în|-in/ | /-aye/ | /hebûynaye/ |
| pr. perfect | Ind. | /he-|ne-/ | /-ibû-/ | /-im|-ît|-e|-în|-in/ | - | /hebûyn/ |
| pr. perfect | Subj. | /he-|ne-/ | /-ibûbêt-/ | /-im|-ît|Ø|-în|-in/ | - | /hebûbêtîn/ |

Table 12: Morphotactics of the verbs /hebûn/ 'to be, to exist'

- Having izafe clitic but indeclinable: /sereřa-y/ '', /le-dwa-y/ '', /be-gwêre-y/ '', /be-pê-y/ '', /be-ho-y/ '', /le-řê-y/ '', /derbare-y/ '', /lewete-y/ '', /be-mebest-î/ '' and so on.
- The indeclinable compound preposition /sebaret be/ is written as two words.

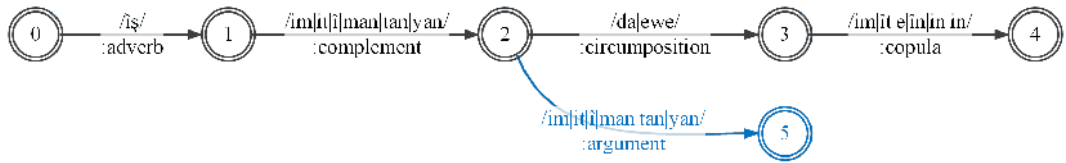

Figure 5: The morphotactics of declinable prepositions. Only /pê/, /lê/, /tê/, and /bo/ can take the argument clitic.

Figure 5 shows the morphotactics of prepositions. The subject/object clitic occurs only with /pê/, /lê/, /tê/ and /bo/ prepositions. Prepositions /pê/ and /lê/ are declinable equivalents of /be/ and /le/ respectively; e.g., /pê-m-tan gut/=/be min-tan gut/ 'you told me' and /lê-m-î sendewe/=/le min-î sendewe/ 'he took it back from me'.

The circumpositions in CK are: /be ...-da/ 'through', /le ...-da/ 'in', /legeł ...-da/ 'with', /be ...-ewe/ 'with', /le ...-ewe/ 'from', /bo ...-ewe/ 'toward'. The second part of the circumposition attaches to its complements; e.g., in /le Hewlêr-ewe/ 'from Arbil'.

The postverb /-e/ occur as a replacement of prepositions /be/ or /bo/.

## 4.6 Adverbs and Other Word Classes

In CK, adverbs, interjections, conjunctions and the other particles are indeclinable. The adverbs may only receive displaced clitics of /-îş/ and /im|it|î|man|tan|yan/ (McCarus, 1958, p. 71). Infrequently and in non-standard structures, conjunctions accept /-îş/; e.g., /hem-îş/.

Adjectives in their uninflected form can be converted into adverbs with zero derivation (MacKenzie, 1961, §189). For example, /ʔazayane/ in /ʔazayane biřyarim da/ 'I decided bravely' is an adverb but in /ʔazayane-tirîn biřyar/ 'bravest decision' is an adjective. Also, uninflected nouns can be used as adverbs. For example, /ʔemsał/ in /ʔemsał deřoyn/ 'we will go this year' is an adverb but in /budce-y ʔemsał/ 'this year's budget' is a noun.

The compound adverbs are frequent; e.g., /be-dax-ewe/ 'unfortunately', /be-aram-î/ 'unfortunately'.

The following represent examples:

- Adverbs: /ʔêre/ 'here', /ʔêsta/ 'now', /hergîz/ 'never', /be-taybet/ 'especially', /zor/ 'very'
- Interjections: /ʔaferîn/ 'bravo', /dek/ 'oh', and /dirêx̌/ 'alas'
- Conjunctions: /bełam/ 'but', /hem/ 'too', and /ʔeger/ 'if'.
- Particles: /ʔa/ 'yes', /ba/ 'yes (why not)', /bełê/ 'yes', /na/ 'no'.

## 4.7 Derivation

In this study, we review more generative and rule-based derivational affixes to reduce the amount of data stored in the lexicon. The reduction of lexical redundancy facilitates the maintenance of the analyzer. Some derivational suffixes have limited generativity; for example, suffix /-bar/ forms only limited well-formed words such as /tawan-bar/ 'criminal' and /xefet-bar/ 'sorrowful'. However, in structures of infinitive, past participle and verbal compounds, the affixes generate new words with all corresponding stem classes. These structures are elaborated on in the following.

| Tense/Aspect | Mood | 0 | +1 | +2 | +3 | +4 | Example |
|---|---|---|---|---|---|---|---|
| present | Ind. | /he-\|nî-/ | (clitic) | - | /-im\|-ît\|-e\|-în\|-in/ | - | /he-man-e/ /...-man he-ye/ |
| present | Subj. | /he-\|ne-/ | (clitic) | /-ib/ | /-im\|-ît\|-êt\|-în\|-in/ | - | /he-man-bêt/ /...-man he-bêt/ |
| simple past | Ind. | /he-\|ne-/ | (clitic) | /-ibû-/ | /-im\|-ît\|Ø\|-în\|-in/ | - | /he-man-bû/ /...-man he-bû/ |
| simple past | Subj. | /he-\|ne-/ | (clitic) | /-ibû-/ | /-im\|-ît\|Ø\|-în\|-in/ | /-aye/ | /he-man-buw-aye/ /...-man he-buw-aye/ |
| pr. perfect | Ind. | /he-\|ne-/ | (clitic) | /-ibû-/ | /-im\|-ît\|-e\|-în\|-in/ | - | /he-man-buw-e/ /...-man he-buw-e/ |
| pr. perfect | Subj. | /he-\|ne-/ | (clitic) | /-ibûbêt-/ | /-im\|-ît\|Ø\|-în\|-in/ | - | /he-man-bûbêt/ /...-man he-bûbêt/ |

Table 13: Inflection of the verbs /hebûn/ 'to have'

| Person/Number | To be an adjective | To have a quality |
|---|---|---|
| 1SG | /birsî-m/ | /birsî-m-e/ |
| 2SG | /birsî-yt/ | /birsî-t-e/ |
| 3SG | /birsî-ye/ | /birsî-ye-tî/ |
| 1PL | /birsî-yn/ | /birsî-man-e/ |
| 2PL | /birsî-n/ | /birsî-tan-e/ |
| 3PL | /birsî-n/ | /birsî-yan-e/ |

Table 14: The difference between to be an adjective and to have a quality.

### 4.7.1 Infinitive

In CK, an infinitive is a verbal noun made from past stem of the verb and the suffix /-in/; e.g., /girt-in/ 'to take'. Furthermore, causative and passive infinitives are made from corresponding past stems; e.g., /gîra-n/ 'taking' (passive of /girtin/) and /şik-and-in/ 'breaking' (passive of /şika-n/). The negative infinitive is made by adding the prefix /ne-/; e.g., /ne-girt-in/ 'not taking'. The meaning of infinitives change with preverbs, and postverb /-ewe/; e.g., /heł-girt-in/ 'lifting up', /girt-in-ewe/ 'taking back'.

### 4.7.2 Past Participle

A past participle is an adjective made up from past stem and suffix /-û/ (/w/ after vowels), e.g., /hat-û/ 'came, received' from /hat-in/ 'to come'. New meanings are formed by affixation of negative prefix (/ne-hat-û/'not came') and preverbs (/heł-hat-û/ 'escapee' and /heł-ne-hat-û/ 'not escapee'). The separate past participle is well-formed only for passive and some intransitive verbs. For example, from transitive verb /xwardin/ 'to eat', separate /*xward-û/ is ill-formed but compound /peng-xward-û/ '' is well-formed.

### 4.7.3 Verbal Compounds

In CK, the verbal compounds are made up from a non-verbal part (noun, adjective, adverb, preposition, and their combination) and a verbal part (McCarus, 1958). In SCK orthography, constituents of compound verbs are written separately but compound nouns and adjectives are written as one word; e.g., the infinitive /taqet-girtin/ (تاقەتگرتن) 'being patient' and the adjective /taqet-gir/ تاقەتگر 'patient' are considered as one word but the conjugated verb /taqet bi-gir-e/ (تاقەت بگرە) 'be patient!' is considered as two words. Therefore, in the analyzer we have to analyze the compound nouns and adjectives.

The structures of compound past participle and compound agent noun/adjective depend on the transitivity of the verbal part and the final compound. The compound infinitive, like simple ones, are made by adding the suffix /-in/ and varying by adding prefix /ne-/ (negation) and postverb /-ewe/. Table 15 shows derived words from compound verbs with examples.

| Transitivity | Infinitive | Past participle | Agent |
|---|---|---|---|
| transitive (transitive verbal) | /dwa(-ne)-xistin(-ewe)/ /dwa(-ne)-xiran(-ewe)/ (passive) | /dwa(-ne)-xiraw(-tir/tirîn)/ (from passive stem) | /dwa(-ne)-xer(-tir/tirîn)/ (from present stem) |
| intransitive (transitive verbal) | /swênd(-ne)-xwardin(-ewe)/ | /swênd(-ne)-xwardû(-tir/tirîn)/ (from past stem) | /swênd(-ne)-xor(-tir/tirîn)/ (from present stem) |
| intransitive (intransitive verbal) | /tê(-ne)-geyştin(-ewe)/ /tê(-ne)-geyandin(-ewe)/ (causative) | /tê(-ne)-geyştû(-tir/tirîn)/ (from past stem) | /tê(-ne)-geyên(-tir/tirîn)/ (from causative stem) |

Table 15: Examples of words derived from compound verbs.

The "incorporated verbal compound", as /nan-xwardin/, /masî-girtin/, /man-girtin/ are common and generative. They are intransitive and they have compositional meaning (Karami, 2017). The verb /dest-biřîn/ can be both a compound verb as */fêłbazêk destî biřîyn/* 'a scammer deceived us' or an incorporated compound as */çeqoke destî biřîyn/* 'The knife cut our hand'.

All Adjectives can form compounds with verbs */kirdin/* 'to do' and */bûn/* 'to be'. For example, with adjective */cwan/* 'beautiful', we can generate various compound words:

- */cwan(-tir)(-ne)-bûn(-ewe)/* (intransitive infinitive),
- */cwan(-tir)(-ne)-kirdin(-ewe)/* (transitive infinitive),
- */cwan(-tir)(-ne)-kiran(-ewe)/* (passive infinitive),
- */cwan(-tir)(-ne)-kira-w/* (past-participle),
- */cwan(-tir)(-ne)-ker)/* (agent name).

The reflexive pronoun */xo-/* combines with many transitive verb stems and forms compound infinitives, such as */xo-kuştin/* 'to commit suicide', */xo-dizînewe/* 'to skive', */xo-nizîk-kirdinewe/* 'to bring yourself closer'.

## 4.8 Contractions

Due to the high frequency of the two preposition */le/* and */be/* before demonstrative adjectives/pronouns, */ʔem/* 'this' and */ʔew/* 'that', their contractions (*/bem/*</*/be ʔem/*, */bew/*</*/be ʔew/*, */lem/*</*/le ʔem/*, and */lew/*</*/le ʔew/*) are common and frequent. The morphotactics of the resulting contraction is the same as the final part; e.g., */bem-an-im gut/* 'I told them' and */lem kiçane/* 'from these girls'. The contractions of these prepositions preceding the adverbs */ʔêre/* 'here' and */ʔewê/* 'there' are also common: */lêre/*</*/le ʔêre/*, */lewê/*</*/le ʔewê/*, */bêre/*</*/be ʔêre/*, */bewê/*</*/be/ /ʔewê/*. Table 16 shows the contractions with prepositions */le/* and */be/*.

|           | /ʔem/ 'this' | /ʔew/ 'that' | /ʔêre/ 'here' | /ʔewê/ 'there' |
|-----------|--------------|--------------|---------------|----------------|
| **/le/** 'from'     | /lem/        | /lew/        | /lêre/        | /lewê/         |
| **/be/** 'with, to' | /bem/        | /bew/        | /bêre/        | /bewê/         |

Table 16: Contractions with prepositions */le/* and */be/*.

With numbers */dû/* 'two' and */sê/* 'three', the following contractions are common: */duwan/*</*/dû dane/* 'two pieces', */syan/*</*/sê dane/* 'three pieces', */herdûkyan/*</*/her dû-eke-yan/* 'both of them', and */hersêkyan/*</*/her sê-eke-yan/* 'each three of them'. With */yek/* 'one' the contractions */pêk/*</*/be-yek/* and */lêk/*</*/le-yek/* are frequent, especially in compounds; e.g., */pêk-da-dan/* < */be-yek-da-dan/*.

## 4.9 Semantic Restrictions

To prevent the generation of ill-formed words, we have taken into account the following semantic restrictions in our implementation:

- Some nouns, such as */ʔêre/* 'here, this place' and */ʔewê/* 'there, that place', */ʔemsał/* 'this year' are semantically definite and do not accept definition, plural, demonstrative and possessive suffixes.

- In transitive verbs, the subject and object pronouns cannot be both in first person or second person. For example /*de-m-nas-im/ '*I know me' and /*de-tan-nas-ît/ '*you know you' are ill-formed. In these situations, reflexive pronouns are used: for example, */xo-m de-nas-im/* 'I know myself'.

- Some transitive verbs like */gutin/* (*/witin/*) 'to say', */zanîn/* 'to know' and */wêran/* 'to dare' only accept third person direct objects. Similarly, their passive verbs and some intransitive verbs like */ʔêşan/* 'to ache' only accept third person subject. For example, /*de-man-gut-in/ and /*de-zanr-a-yn/ are ill-formed.

- Among the inflectional forms of reflexive pronoun */xo-/*, combinations of /*xo-m-man/, /*xo-t-tan/, and /*xo-y-yan/ are ill-formed.

- Reciprocal pronoun */yektir/* 'each other' occurs only with plural arguments, therefore it does not accept singular clitics, e.g., /*yektir-im/ is not acceptable.

# 5 Orthographic Rules

In this section, we describe the orthographic rules of CK. These rules govern the changes that occur in the orthography of words in their morpheme junctures. In general when the pronunciation hardens by adjoining the morphemes in a word, a morphophonemic change occurs to ease the pronunciation. An orthographic rule describes a morphophonemic change that becomes visible in the orthography. In the orthographic system of CK, which is adapted from Arabic script, there are three cases where letters and the phonemes do not have one-to-one mapping (Mahmudi & Veisi, 2021). The letter "ی" represents both */y/* and */î/*, letter "و" represents */w/*, */u/*, and */û/*, and there is no letter for vowel */i/*. These cases hide some morphophonemic changes in written form. For example, if vowel /î/ in 2SG pronoun suffix /-ît/ proceed a vowel, it turns into approximant /y/, as in the word "گەورەیت" /gewre-yt/ 'you are big'. However, the CK orthography does not show this morphophonemic change as "ی" represents both /y/ and /î/.

We extracted the morphophonemic changes from the descriptive linguistic resources such as MacKenzie (1961), McCarus (1958), and Mohammadi (2014). However, as some rules may vary in different dialects of CK, we have focused generally on SCK, by evaluating rules with official news corpora, such as AsoSoft Corpus (Veisi et al., 2019). In SCk, morphophonemic changes occur when two vowels adjoin at the morpheme juncture. Therefore, we focus on the bound morphemes starting or ending in a vowel. We have summarized their behavior in the following tables: Table 17 shows the SCK morphophonemic changes which are visible in the orthography for each bound morpheme, and Table 18 shows morphophonemic changes that do not become visible in the CK's orthography.

| Suffix | Functions | Allomorphs | Examples |
| --- | --- | --- | --- |
| /-û/ | past participle | /-w/ | /xura-w/ |
| /-û-/ | present perfect | /-w-/ | /kiřî-w-e/ |
| /-aye/ | perfect and imperfect subjunctive | /-ye/ | /bikira-ye/ |
| /-e/ | simple past | /-ye/ | /kira-ye/ |
| /-e/ | postverb in present perfect and subjunctive imperfect | /-te/ | /hatuwe-te/ /bihataye-te/ |
| /-ewe/ | postverb in simple past | /-yewe/ | /kira-yewe/ |
| /-ewe/ | postverb in imperatives | /-rewe/ | /bipirs-e-rewe/ |
| /-ewe/ | postverb in present perfect and subjunctive imperfect | /-tewe/ | /hatuwe-tewe/ /bihataye-tewe/ |
| /-îş/ | adverbial clitic | /-ş/ | /wata-ş/ |
| /-em/ | ordinal numbers | /-yem/ [/-hem/] | /sê-yem/ |
| /-emîn/ | ordinal numbers | /-yemîn/ [/-hemîn/] | /sê-yemîn/ |
| /-e/ | 2SG pronoun in imperatives | Ø | /biłê-Ø/ |
| /-êt/ | 3SG pronoun in present verbs | see Table 20 | |
| /-ewe/ | circumposition | see Table 19 | |
| /-êk/ | indefinite | see Table 19 | |
| /-eke/ | definite | see Table 19 | |
| /-an/ | plural | see Table 19 | |
| /-e/ | definite izafe | see Table 19 | |
| /-e/ | demonstrative | see Table 19 | |

Table 17: The morphophonemic changes visible in the orthography (non-standard allomorphs in brackets).

| Suffix | Function | Allomorphs | Examples |
| --- | --- | --- | --- |
| /-û-/ | present perfect | /-uw-/ | /hat-uw-e/ |
| /-ibû/ | present perfect | /-bû/ | /kira-bû/ |
| | past perfect | /-ibuw/ | /hat-ibuw-e/ |
| | past perfect | /-buw/ | /kira-buw-e/ |
| /-îş/ | adverbial clitic | /-yş/ | /wata-ş/ |
| /-î/ | izafe clitic | /-y/ | /wata-y/ |
| /-im/ | 1SG possessive pronoun | /-m/ | /wata-m/ |
| /-it/ | 2SG possessive pronoun | /-t/ | /wata-t/ |
| /-î/ | 3SG possessive pronoun | /-y/ | /wata-y/ |
| /-în/ | 1PL | /-yn/ | /hatû-yn/ /wata-yn/ |
| /-in/ | 2PL/3PL | /-n/ | /hatû-n/ /wata-n/ |
| /-ît/ | 2SG | /-yt/ | /hatû-yt/ /wata-yt/ |

Table 18: The morphophonemic changes that are not visible in the orthography

Since the behavior of some morphophonemic changes depends on the final phoneme of the preceding morpheme, we present them in separate tables. Table 19, with examples, shows the orthographic changes occurring with the bound morphemes /-ewe/, /-eke/, /-an/ and /-e/; and, Table 20 shows the morphophonemic changes occurring with 3SG forms of present verbs. It should be noted that in the present perfect verbs, if the verb's past stem ends in /û/ vowel (as /bû/ 'was/were', /çû/ 'went', /dirû/ 'sewed'), the final /û/ of the stems becomes /u/; e.g., /çuwim/ 'I have gone'. Also, in the ordinal forms of /dû/ 'two', /û/ vowel of the stem changes; e.g., /duwem/ and /duwemîn/.

Some linguists, like (MacKenzie, 1961), consider diphthong /ö/ [œɛ] a phoneme of CK. In the CK orthography, it is shown as digraph وێ. Table 19 shows the only clue for this claim, as the morphophonemic change of definite suffix for words ending in diphthong /wê/ is different from words that end in /ê/.

| Stem's final | Stem | Definite | Indefinite | Plural | izafe | demonstrative | discont. preposition |
|---|---|---|---|---|---|---|---|
| /w/ | /pyaw/ | /pyaw-eke/ | /pyaw-êk/ | /pyaw-an/ | /pyaw-e gewreke/ | /ew pyaw-e/ | /be pyaw-ewe/ |
| /y/ | /mey/ | /mey-eke/ | /mey-êk/ | /mey-an/ | /mey-e gewreke/ | /ew mey-e/ | /be mey-ewe/ |
| /û/ | /beřû/ | /beřuw-eke/ | /beřû-yek/ | /beřuw-an/ | /beřuw-e gewreke/ | /ew beřuw-e/ | /be beřuw-ewe/ |
| /î/ | /masî/ | /masî-yeke/ | /masî-yek/ | /masî-yan/ | /masî-ye gewreke/ | /ew masî-ye/ | /be masî-yewe/ |
| /e/ | /ajawe/ | /ajawe-ɵke/ | /ajawe-yek/ | /ajawɵ-an/ | /ajawe-ɵ gewreke/ | /ew ajawe-ye/ | /be ajawe-ɵwe/ |
| /a/ | /çya/ | /çya-ɵke/ | /çya-yek/ | /çya-yan/ | /çya-ɵ gewreke/ | /ew çya-ye/ | /be çya-ɵwe/ |
| /o/ | /diro/ | /diro-ɵke/ | /diro-yek/ | /diro-yan/ | /diro-ɵ gewreke/ | /ew diro-ye/ | /be diro-ɵwe/ |
| /ê/ | /dê/ | /dê-ɵke/ | /dê-yek/ | /dê-yan/ | /dê-ɵ gewreke/ | /ew dê-ye/ | /be dê-ɵwe/ |
| /wê/ | /gwê/ | /gwê-yeke/ | /gwê-yek/ | /gwê-yan/ | /gwê-ɵ gewreke/ | /ew gwê-ye/ | /be gwê-ɵwe/ |

Table 19: The orthographic changes with bound morphemes /-ewe/, /-eke/, /-an/ and /-e/ based on preceding phoneme.

| Stem's final | Changes | Example |
|---|---|---|
| /ê/ | /ê/+/ê/ => /ê/ | /dełê/+/êt/ => /dełêt/ |
| /e/ | /e/+/ê/ => /a/ | /deke/+/êt/ => /dekat/ |
| /o/ | /o/+/ê/ => /wa/ | /dexo/+/êt/ => /dexwat/ |
| /û/ | /û/+/ê/ => /uwê/ | /denû/+/êt/ => /denuwêt/ |
| /î/ | /î/+/ê/ => /yê/ | /dejî/+/êt/ => /dejyêt/ |

Table 20: The morphophonemic changes in third singular of present verbs.

## 5.1 Exceptional Morphophonemic Rules

The exceptional morphophonemic rules occur only in particular paradigms of certain verbs. In SCK, the following exception occur:

- In past verbs of /heł-stan/ 'to stand up', the form /heł-sta/ changes to /hesta/; e.g., /hesta/ 'he stood up';
- In imperative forms of /heł-stan/ the combination /hełbist/ becomes /hest/ to forming /hestin/ 'get up!';
- In past verbal forms of /heł-hatin/ 'to run away' the /heł-hat/ turns into /hełat/ 'he ran away';
- In the 2SG imperative form of the verb /çun/ 'to go' the pronoun /-e/ becomes /-o/: /biço/ 'go!';
- As noted earlier, the present tenses of /wîstin/ 'to want' are irregular in getting pronominal clitics as the subject. When the subject pronoun has moved to an earlier position in the utterance, as /min to-m dewê/ 'I want you', the initial vowel of the stem drops for preventing hiatus between the stem /-ewê/ and the prefixes; e.g., /dewê/ </de-/+/-ewê/ and /nawê/ </na-/+/-ewê/; and,
- Present and imperative paradigms of /hatin/ 'to come', as described earlier. Note that all present and imperative prefixes (/de/, /na/, /bi/, /ne/, /me/) ends in vowel and as the present stem of the verb /hatin/ 'to come' is only a vowel (/ê/), morphophonemic changes occur for preventing hiatus in both sides of the stem. Table 21 shows all paradigms of the verb /hatin/ based on present stem.

| Verb form | 1SG | 2SG | 3SG | 1PL | 2PL | 3PL |
|---|---|---|---|---|---|---|
| positive indicative | /dêm/ | /dêyt/ | /dêt/ | /dêyn/ | /dên/ | /dên/ |
| positive subjunctive | /bêm/ | /bêyt/ | /bêt/ | /bêyn/ | /bên/ | /bên/ |
| negative indicative | /nayem/ | /nayeyt/ | /nayet/ | /nayeyn/ | /nayen/ | /nayen/ |
| negative subjunctive | /neyem/ | /neyeyt/ | /neyet/ | /neyeyn/ | /neyen/ | /neyen/ |
| positive imperative | - | /were/ | - | - | /werin/ | - |
| negative imperative | - | /meye/ | - | - | /meyen/ | - |

Table 21: Paradigms of present tenses of the verbs /hatin/ 'to come'.

## 5.2 Non-standard Orthographic Changes

Here, we draw upon orthographic changes that occur in non-standard CK texts. Note that the more common change involves dropping final /t/ from bound morphemes.

When 2SG pronoun /-ît/ is word-final, its /t/ is usually dropped; e.g., /hat-î(t)/ 'you came' but in non-final positions such as /hat-ît-ewe/ 'you came again' /t/, it remains. In northern dialects of CK like Mukri (Kalbasi, 1983) and Arbili dialects (MacKenzie, 1961), 2SG pronoun morpheme is /-î (-y)/ both in final and non-final positions (/hat-î/ 'you came' and /hat-î-yewe/ 'you came again'). Similarly, when bound morpheme /-êt/ (3SG pronoun in present verbs) is word-final, its /t/ is usually dropped in non-standard texts; e.g., /de-gir-ê(t)/ 'he takes' but in non-final position, its /t/ is preserved as in /de-gir-êt-ewe/ 'he takes again'. In present perfect verbs, in non-formal texts, sometimes /-uwete/ becomes /-ote/. For example, /hat-ote/ instead of standard /hat-uwete/ 'he has came to ...' and /hat-otewe/ instead of standard /hat-uwetewe/ 'he has came back'. In northern dialects of CK, the orthographic changes are also different for postverbal clitics /-ewe/ and /-e/. In SCK, if they occur after a vowel, a euphonic consonant /r/ is inserted, but in Mukri and Arbili, the vowel of the morpheme drops. For example, the standard form /bîxe-re/ 'throw it to ...!' and /mexo-rewe/ 'do not drink!' become /bîxe-Ø/ and /mexo-we/, respectively. In some dialects, the initial /ye/ (or /he/) of ordinal suffix is dropped (MacKenzie, 1961); e.g., /heşta-min/ 'eightieth' compared to standard form of /heşta-yemin/. Table 22 lists three common morphophonemic changes in non-standard CK texts that occur in prefix of verbs whose stem's initial phoneme is /h/ or /ʔ/.

| Non-standard Form | Standard Form | Examples |
| --- | --- | --- |
| /dê-/ (/yê/) | /de-hê-/ /de-ʔê-/ | /dênim/ < /dehênim/ 'I bring' /dêłim/ < /dehêłim/ 'I let' /dêşêt/ < /dêʔêşêt/ 'it pains' |
| /bê-/ | /bi-hê-/ /bi-ʔê-/ | /bênim/ < /bihênim/ 'may I bring' /bêłim/ < /bihêłim/ 'may I let' /bêşêt/ < /biʔêşêt/ 'may it pains' |
| /naye-/ | /na-hê-/ /na-ʔê-/ | /nayenim/ < /nahênim/ 'I do not bring' /nayełim/ < /nahêłim/ 'I do not let' /nayeşêt/ < /naʔêşêt/ 'it does not pain' |
| /eye-/ | /e-hê-/ /e-ʔê-/ | /meyenin/ < /mehênin/ 'do not bring!' /neyełim/ < /nehêłim/ 'if I do not let' /neyeşêt/ < /neʔêşêt/ 'if it does not pain' |

Table 22: The common non-standard morphophonemic changes with verbs whose stem's initial phoneme is /h/ or /ʔ/.

# 6 Implementation

In this section, the implementation process of our morphological analyzer is described. We hope that it will be useful for other morphologically complex and low-resourced languages, especially the members of the Kurdish language family. The affix-stripping tools such as Hunspell [1] were tried for the implementation. However, the CK morphology was too complex to be handled properly by such tools. Instead, we used two-level morphology with finite state transducers (FSTs). The advantages of using FST tools include the following: all complex features of the CK Morphology are implementable; the system is reversible, i.e., it can be used for both analysis and generation; and, the maintenance of the lexicon is easier and the lexicon is more generative.

## 6.1 Procedure

We have used HFST (Lindén & Pirinen, 2009) which is an open-source toolkit that supports the implementation of simple and weighted finite state transducers. This framework supports Unicode encoding and can be used for developing morphological analyzers (Lindén et al., 2011). Also, it can be leveraged in other language processing applications such as spell-checkers (Lindén et al., 2013). Since the specification of weights needs statistical resources about the use of word stem and bound morphemes and these data are not available for CK, in our proposed analyzer, we have used simple (non-weighted) finite state transducers. In the HFST framework, the lexicon and the morphotactic rules are coded as one LEXC file, and the orthographic (morphophonemic) rules are coded as a TWOL file. The details about LEXC and TWOL scripts are found in Beesley and Karttunen (2003). We have used the standard alphabet of CK (Right-to-left, based on Arabic script) with Unicode characters specified by Department of IT of Kurdistan Regional Government (2014).

The overall process of creating CKMorph is as follows:

- *Minimal lexicon preparation*: First, we prepare a minimal lexicon with several words that have various features in each stem class.
- *Coding LEXC and TWOL*: Then, we code the LEXC and TWOL files according to the morphotactic rules presented in section 4 and the orthographic rules presented in section 5.
- *Compiling FST*: The next step is to compile the FST for checking the analyzer with a list of inflected samples selected by authors (native speakers) from the context inside the corpus.
- *Correcting LEXC and TWOL files*: If any errors are seen in analyses, we revise the LEXC and TWOL files.
- *New stems discovery*. Finally, new stems are added to their classes in the lexicon, all checked manually according to the dictionaries and their uses inside the corpus.

---

[1] http://hunspell.github.io/

## 6.2 Challenges in Implementation

Since the standardization process of Kurdish has started in the 20th century (Hassanpour, 1992), the non-standard forms can still be seen abundantly in the written texts. In CKMorph, we have compiled the non-standard stems and morphemes with special tags that help the maintenance of the lexicon for future applications like spell-checkers. These non-standard forms are either lexical or morphological. The lexical variations found in the corpora are primarily the loanwords and especially the proper nouns. For example, the Arabic given name "Muhammad" is written in CK as موحەممەد, موحەمەد, محەمەد, محەممەد and محمد. For such cases, we generally have chosen the most frequent variation (according to the AsoSoft corpus (Veisi et al., 2019)) as the standard form and the others are tagged as the non-standard forms. The Kurdish-origin words are also recorded differently in the dictionaries, mainly due to dialectal variation. Finding all common variations of a word and selecting the most frequent one was a time-consuming task.

The morphological variations in CK morphotactics are divided into two categories: dialectal and free variations. Using different bound morphemes and sometimes with different order indicates the dialect of the author. For example, the progressive (imperfect) prefix in SCK is */de/* and in the southern dialects of CK, it is */e-/*. The free variations of some bound morphemes especially in word-final positions are common. For example, the indefinite suffix */-êk/* and the 2SG marker */-ît/* have another form in word-final position with dropping the final stop consonant (*/-ê/* and */-î/*, respectively). We have implemented common morphological variations with a non-standard tag in the underlying level.

A crucial challenge in implementation of CK morphotactics is management of the numerous word patterns that include a verb stem. In these patterns, the verbal stem, the prefixes and the suffixes together form the word and any misarrangement causes an ill-formed word. In SCK, there are hundreds of simple verb lemmas. Each one has three stems (present, past, and causative/passive) and each stem can be used in the formation of various types of words. Consequently, each word, inflected or derived from these verb stems, follows a different path in the FSTs. Additionally, to prevent illegal paths in the transducer, we should consider the valency and the semantic limitations of each verb and the far dependencies of the bound morphemes in the word-formation patterns.

A solution to deal with this challenge in the LEXC file is repeating the verb stems for each pattern type. However, the repetition of all the hundreds of the verb stems is not efficient and it makes the maintenance unmanageable. Instead, we use "flag diacritics" which are treated like epsilons in FSTs and do not appear in the output. In fact, they act like instructive memories to remove illegal paths (Beesley & Karttunen, 2003). We need to use various flag diacritics in the LEXC file. For example, In the first prefix of verbal patterns, we specify the pattern type using @U.Word.vi@, which is a unification flag diacritic meaning the word should follow the patterns of an intransitive verb. In the next continuation lexicons of the LEXC file, the formation of the word is proceeded in the corresponding pattern using the "require" and "disallow" flags (@R.Word.vi@ and @D.Word.vi@, respectively).

Figure 6 shows a simplified LEXC file for implementing some CK word-formation patterns for generating these words:

- وەرین:وەریم[VERB]{1SG}
- داوەرین:داوەریم[VERB]{1SG}
- وەرین:وەرین[NOUN]
- داوەرین:داوەرین[NOUN]
- وەرین:وەرینێک[NOUN]{indefinite}
- داوەرین:داوەرینێک[NOUN]{indefinite}

Another use of flag diacritics is in the semantic restrictions with the preverbs. For example, since the lemma */xizan/* 'to slip' can only take preverbs */heł/* and */da/*, we use "D flags" to disallow the other preverbs (e.g., @D.Preverb.Ra@ for preventing */řa-/* preverb).

In summary, the flag diacritics are used in our analyzer for the following situations:

- Leading the word-formation patterns in words including a verb stem;
- Controlling the far dependency of bound morphemes in verb with multiple affixes;
- Preventing semantically illegal pronouns, such as */*xward-man-im/* or */*de-tan-xward-ît/*;
- Controlling the preverb valency of each verb; i.e., can the verb take */heł-/, /da-/, /řa-/, /řo-/, /wer-/, /ser-/*, or */der-/* or can it occur without preverb?;
- Controlling valency of verbs to take post-positions */-e/* and */-ê/*. For example, */geyiştim-ê/* is valid and */*kewtim-ê/* is ill-formed;
- Preventing the */îş/* clitic and pronominal clitics, if a previous displaced position is filled;
- Controlling the necessity of epenthesis (euphonic) between adjacent vowels (*/-t-/* in present perfect and */-r-/* in imperatives);
- Controlling the formation of free past participle. For example, */hat-û/* is a valid word but */*xward-û/* is illegal if it occurs separately. Only the transitive compounds can form a free past participle;
- Controlling the necessity of */-ewe/* for verbs such as */şardin-ewe/*;
- Preventing the direct object marker in certain transitive verbs; and,

```
 1  Multichar_Symbols
 2  ! flag diacritics:
 3  @P.PreVerb.da@ @P.PreVerb.wer@ @D.PreVerb.wer@
 4  @U.Word.inf@ @R.Word.inf@ @U.Word.vi@ @R.Word.vi@
 5  ! POS tags and morphological features:
 6  [VERB] [NOUN] {1SG} {indefinite}
 7
 8  LEXICON Root
 9  @U.Word.inf@      PreVerb;
10  @U.Word.vi@       PreVerb;
11
12  LEXICON PreVerb
13  VerbStems;
14  @P.PreVerb.da@د‌ا       VerbStems;
15  @P.PreVerb.wer@وەر      VerbStems;
16
17  LEXICON VerbStems
18  @D.PreVerb.wer@ وەرینv;   ! disallow وەر preverb for verb وەرین
19
20  LEXICON وەرینv
21  وەری         ViPast;
22  وەر:وەری     ViPresent;
23  وەر:وەری     Causative;
24
25  LEXICON ViPast
26  @R.Word.vi@ن[VERB]:@R.Word.vi@      ViPast+1;
27  @R.Word.inf@ن[NOUN]:@R.Word.inf@ن   N+1;
28
29  LEXICON ViPast+1
30  {1SG}:م      #;
31
32  LEXICON N+1
33  #;
34  {indefinite}:یەک   #;
```

Figure 6: A simplified LEXC file for implementing some CK word-formation patterns, with our custom highlighting in Notepad++.

- Preventing the free occurrence of bound stems. For example, the stem /he-/ in verb /hebun/ is a bound morpheme.

As noted in the previous sections, there are several exceptions in the CK morphotactics and orthographic rules. For example, the imperative forms of the verb /hatin/ 'to come' are exceptional as they are made up from stem /wer/ instead of the present stem /ê/ (e.g., /wer-e/ 'come!', /wer-in-e êre/ 'come here!') and they do not take prefix /bi-/ or clitic /îş/. We implement numerous rules in the LEXC and TWOL files to handle these exceptions. The flag diacritics are also used to prevent the illegal paths in the FSTs.

### 6.3 Covering Common Spelling Errors

The spelling error occur in all texts. For broader coverage of the analyzer, we added some rules to the morphotactics to support the common systematic spelling mistakes. A frequent spelling problem occurs with the conjunction /û (w)/ 'and'. Although in the normal speech of CK, it is pronounced as an enclitic for the preceding word, it is written as a separated word in standard orthography of CK. However, some writers prefer to type it joined to the previous word, especially when the previous word ends in a non-joining letter (ا د ر ز ژ و ۆ ە). For example, /şar û gund/ 'city and village' is written as شاروگوند. We added an optional و enclitic to the morphotactics of all words (except for particles and conjunctions). Another recurring spelling error in CK texts is typing the prepositions بە and لە to joined to their following word. These rules are specified with a "non-standard tag" for future uses of the analyzer such as spell-checkers.

## 7 Evaluations

### 7.1 Test-set Creation

To evaluate the performance of a morphological analyzer, a gold standard word list with manually checked analyses is needed. There are two options for preparing such a test set: giving all the possible analyses for each word, or giving one analysis according to the context. We opted for the latter option as it can be useful for future tasks such as morphological disambiguation.

In the Unimorph project's repository, a data-set including 22,990 CK words (with one analysis for each word) is freely available[2]. Its words are generated from 274 lemmas and tagged with 34 morphological features. Unfortunately, this word list contains abundant orthographic and morphological errors and its manual correction was not feasible. Instead, we generated our test set by selecting 1,000 random and unique words with their context from the AsoSoft text corpus (Veisi et al., 2019).

---

[2]https://github.com/unimorph/ckb

If the random word was not a CK word, we replaced it by picking up another random one. Also, since we were not evaluating spelling correction, we corrected the spelling mistakes. Then, the authors (all native speakers of Kurdish) manually checked the selected words within the given context, and they determined one correct morphological analysis for each word. This test set is named "CKMorph-Accuracy-TestSet".

To evaluate the coverage of a morphological analyzer, it should be tested on real world texts of a general corpus. For this reason, we evaluate the coverage of CKMorph on the small version of AsoSoft corpus (Veisi et al., 2019)[3]. This subset of AsoSoft corpus contains 5 million tokens collected from CK resources. By purging the punctuation marks, the digits, and the tokens with non-Kurdish characters, we extracted 4,220,806 tokens. A frequency list was prepared to be analyzed by the morphological analyzer. This test set is named "CKMorph-Coverage-TestSet".

We have published our final test sets on Github[4].

## 7.2 Accuracy Evaluation

We analyzed the words of "CKMorph-Accuracy-TestSet" with the latest version of CKMorph analyzer. It generated 3,087 analyses for all words (an average of 3.1 possible analyses for each word; i.e., ambiguity rate of 3.1). Then, we compared the output for each word with the correct golden analysis.

For 95.9% of the test words, the intended analysis was among the analyses that CKMorph generated. Since all tokens of "CKMorph-Accuracy-TestSet" are well-formed words of CK and there are no negative classes in the test set, we can consider this rate as accuracy or recall (true positive rate). The high recall of the analyzer proves that it rarely fails in identifying the well-formed words of the language, and it can be used as the detection system of a spell-checker.

However, in 4.1% of the test words, the analyzer was unable to generate the intended analysis. We can divide the failures into two categories:

1. *Irrelevant analyses*: the analyzer generated some analyses none of which, according to the context, were the intended one. For example, the word "دینا" is a proper noun in "خەباتکار دینا فەرید رادەگەیەنێت" ('activist Dina Farid announces'); however, CKMorph detected it as non-standard forms of "دین+دا" ('in religion') or "دیتن+دا" ('in seeing') instead of the intended proper name.

2. *No analysis*: the analyzer was unable to analyze the word. An example of such words was the word "مەدلوول" ('referent') which is a rarely-used specialized Arabic loanword.

The words whose stems are not added to the lexicon include mainly the named entities, low-frequency words of the language, and non-formal sub-dialectal variations. The unidentified named entities are mainly non-Kurdish names such as "دۆستۆیۆڤسکی" ('Dostoevsky') that have different variations among Kurdish speakers. As noted previously, sub-dialectal variations of CK are abundant and complete coverage is impractical.

Table 23 shows the results of the accuracy evaluation of CKMorph.

| Detection Category | Tokens | Ratio |
| --- | --- | --- |
| Correct analysis | 959 | 95.9% |
| Irrelevant analysis | 5 | 0.5% |
| No analysis | 36 | 3.6% |
| **Total** | **1000** | |

Table 23: Results of the accuracy evaluation of CKMorph.

## 7.3 Coverage Evaluation

We performed the coverage test on "CKMorph-Coverage-TestSet". Without any tokenizations and corrections, the latest version of CKMorph gave at least one analysis for 88.7% of the tokens. Surveying the not-detected tokens showed that a majority of them are common spelling errors that can be divided into two categories:

1. *Common wrong character deletion and replacements*: The deletion of "و/ی" in "یی/وو" clusters, and replacing "ر" with "ڕ" are the common examples.

2. *Multi-word tokens*: Despite orthographic rules of SCK, merging the parts of the compound verbs are common in Kurdish texts. For example, the token "ئەنجامنادەن" should be written as "ئەنجام نادەن" ('they do not perform').

Therefore, we performed corrections on tokens that were not detected by CKMorph. First, we applied one of the several common character insertion/replacement corrections, to see if CKMorph could detect the changed word. For example, inserting a "و" character adjacent to the existing "و" in the token "بوە" gives the well-formed word of "بووە" ('it became'). Then, we split the remaining not-detected tokens into two or three possible segments. For example, by splitting the token "دەستیپێکرد" into three parts, the compound verb "دەستی پێ کرد" ('it started') can be detected by the proposed analyzer.

Table 24 shows our coverage evaluation results.

---

[3] Available at https://github.com/AsoSoft/AsoSoft-Text-Corpus
[4] https://github.com/CKMorph/Evaluations

| Detection | Analyzed Tokens | Coverage |
|---|---|---|
| Detected without corrections | 3,741,995 | 88.7% |
| Detected after common character corrections | 56,216 | 1.3% |
| Detected after one split | 204,339 | 4.8% |
| Detected after two splits | 26,343 | 0.6% |
| Not detected | 191,912 | 4.5% |
| **Total detected** | **4,028,893** | **95.5%** |
| **Total** | **4,220,805** | |

Table 24: Results of the coverage evaluation of CKMorph performed using small version of AsoSoft corpus.

The number of unique words in morphologically complex languages is myriad. The high coverage rate of CKMorph on a general CK corpus with a relatively small lexicon (about 10K lexems), proves that it is possible to cover a vast majority of a morphologically complex language's texts with limited resources.

# 8 Conclusion and Future Works

In this paper, we presented a comprehensive computational model of Central Kurdish (CK) morphology covering all morphological features of the language. We discussed the morphotactics of every CK word category and the orthographic (morphophonemic) rules in the interaction of CK morphemes. We also collected and manually labeled a highly generative lexicon for CK containing all sorts of word stems. We implemented CKMorph Central Kurdish morphological analyzer with the help of these meticulously collected rule sets and resources in a Finite-state transducers framework. Finally, using real-world texts, we prepared test sets for evaluating the accuracy and the coverage of the proposed analyzer. The functionality evaluation of CKMorph indicates its ample coverage of morphologically complex texts of CK, verifying its capability to be a key part of CK processing tools such as spell-checkers, part-of-speech taggers, syntactic parsers, and so forth.

We believe CKMorph will be a valuable resource for downstream applications and future development of similar morphological analyzers. The closely related languages in the same family as CK and especially other dialects of Kurdish share numerous similar features; therefore, Our implementation of CK morphological analyzer can be helpful in accelerating the development of a similar tool the other relative languages. Additionally, in the implementation steps, we have borne the standard forms of the CK in mind. This consideration for standard forms is especially significant as CK as a low-resourced language is in the process of standardization and as of now lacks reliable textual resources. Our effort will facilitate the tedious, time consuming process of preparing such resources, among others. Among other things, this has made feasible developing a spell-checker based on CKMorph with minor modifications.

There are some areas that warrant future lines of research. First, our proposed method provides all possible morphological analyses of the word out of context. This can be followed by a disambiguation phase since a word may have various analyses free from the context. The disambiguation in the situations that the analyzer gives multiple analyses needs a rule-based or data-driven solution. Second, and especially after the disambiguation, the correct analysis can be used in various natural language processing tools such as part-of-speech taggers, lemmatizers, stemmers, syntactic parsers, and text-to-speech applications in the pronunciation of homographs. For instance, developing a spell-checker on the basis of CKMorph can be done as an accessible next work. This will be significant as no spell-checker for CK has been implemented at the time of composition of this paper.

# Data availability

The resources including Central Kurdish verb database (DOI: 10.5281/zenodo.6300522) and evaluation data sets (DOI: 10.5281/zenodo.6300602) are publicly accessible in CKMorph's project's repository at https://github.com/CKMorph.